\pdfoutput=1

\documentclass[11pt]{article}

\usepackage[preprint]{acl}

\usepackage{times}
\usepackage{latexsym}

\usepackage[T1]{fontenc}

\usepackage[utf8]{inputenc}

\usepackage{microtype}

\usepackage{inconsolata}

\usepackage{graphicx}

\usepackage{inconsolata}
\usepackage{enumerate}
\usepackage{enumitem}
\usepackage{tikz}
\usepackage{amsmath}
\usepackage{subcaption}
\usepackage{amssymb}
\usepackage{booktabs}
\usepackage{arydshln}
\usepackage{xcolor}
\definecolor{customorange}{RGB}{237,125,49}
\definecolor{customblue}{RGB}{68,114,196}
\definecolor{customgreen}{RGB}{0,176,80}
\usepackage[normalem]{ulem}
\useunder{\uline}{\ul}{}
\definecolor{royalblue}{rgb}{0.25, 0.41, 0.88}

\usepackage{ulem}
\usepackage{todonotes}

\newcommand{\CPS}{\texttt{ComPSum}}
\newcommand{\AM}{\texttt{AuthorMap}}
\newcommand{\PM}{\texttt{PerMSum}}

\title{Comparative Personalization for Multi-document Summarization}

\author{
  \textbf{Haoyuan Li},
  \textbf{Snigdha Chaturvedi}
\\
  University of North Carolina at Chapel Hill,
\\
  {\{haoyuanl, snigdha\}@cs.unc.edu}
}

\begin{document}
\maketitle
\begin{abstract}
Personalized multi-document summarization (MDS) is essential for meeting individual user preferences of writing style and content focus for summaries. In this paper, we propose that for effective personalization, it is important to identify fine-grained differences between users' preferences by comparing the given user's preferences with other users' preferences. 
Motivated by this, we propose \CPS, a personalized MDS framework. It first generates a structured analysis of a user by comparing their preferences with other users' preferences. The generated structured analysis is then used to guide the generation of personalized summaries. To evaluate the performance of \CPS, we propose \AM, a fine-grained reference-free evaluation framework for personalized MDS. It evaluates the personalization of a system based on the authorship attribution between two personalized summaries generated for different users. For robust evaluation of personalized MDS, we construct \PM, a personalized MDS dataset in the review and news domain. We evaluate the performance of \CPS~on \PM~using \AM, showing that it outperforms strong baselines.
\end{abstract}

\section{Introduction}
\label{intro}
Multi-document summarization (MDS) aims to generate a summary with the salient information from multiple documents on a certain topic, such as multiple news articles about an event \cite{fabbri2019multi} or reviews of a product \cite{bravzinskas2020unsupervised}. However, different users often have different or even conflicting \textit{preferences} of \textit{writing styles} or \textit{content focuses} for summaries \cite{jangpersonalized}. 
While writing style refers to the manner or tone in which the summaries are written, content focus refers to which aspects are emphasized when presenting a certain topic. 
Users can have different preferences for writing style. For example, for product reviews, some users may prefer a formal tone, while others may prefer a conversational tone. User preferences can also differ on \textit{content focus}. Some users may focus on price and utility of the product while others might focus on quality and durability. Therefore, to meet these individual user preferences, personalized MDS is essential. 

Personalized MDS is related to personalized text generation. Recent works on personalized text generation use Large Language models (LLMs) and assume access to the \textit{profile} of individual users--set of documents previously authored by the user. 
They then either retrieve related documents from a user's profile \cite{salemi2024lamp, li2023teach}, include a summary of the user's profile \cite{richardson2023integrating}, or tune different models for different users based on their profiles \cite{tan-etal-2024-democratizing}. 
However, most of these works only include general features of the user and ignore finer differences between users. To identify the finer differences, it is important to compare a user's profile documents with comparable profile documents written by other users. Ideally, the profile documents of two users can be comparable if they are on the same topic but differ on personal preferences. 
In general personalized text generation, identifying such comparable profile documents of different users can be difficult since the differences between profile documents of different users can stem from either personal preferences or topic differences. Contrarily, for MDS, since all input documents are about the same topic (e.g. reviews about the same product), 
 their differences are more likely to stem from differences in personal preferences of their authors (users). 
 
\begin{figure}[t]
\centering
\includegraphics[width=0.48\textwidth, keepaspectratio]{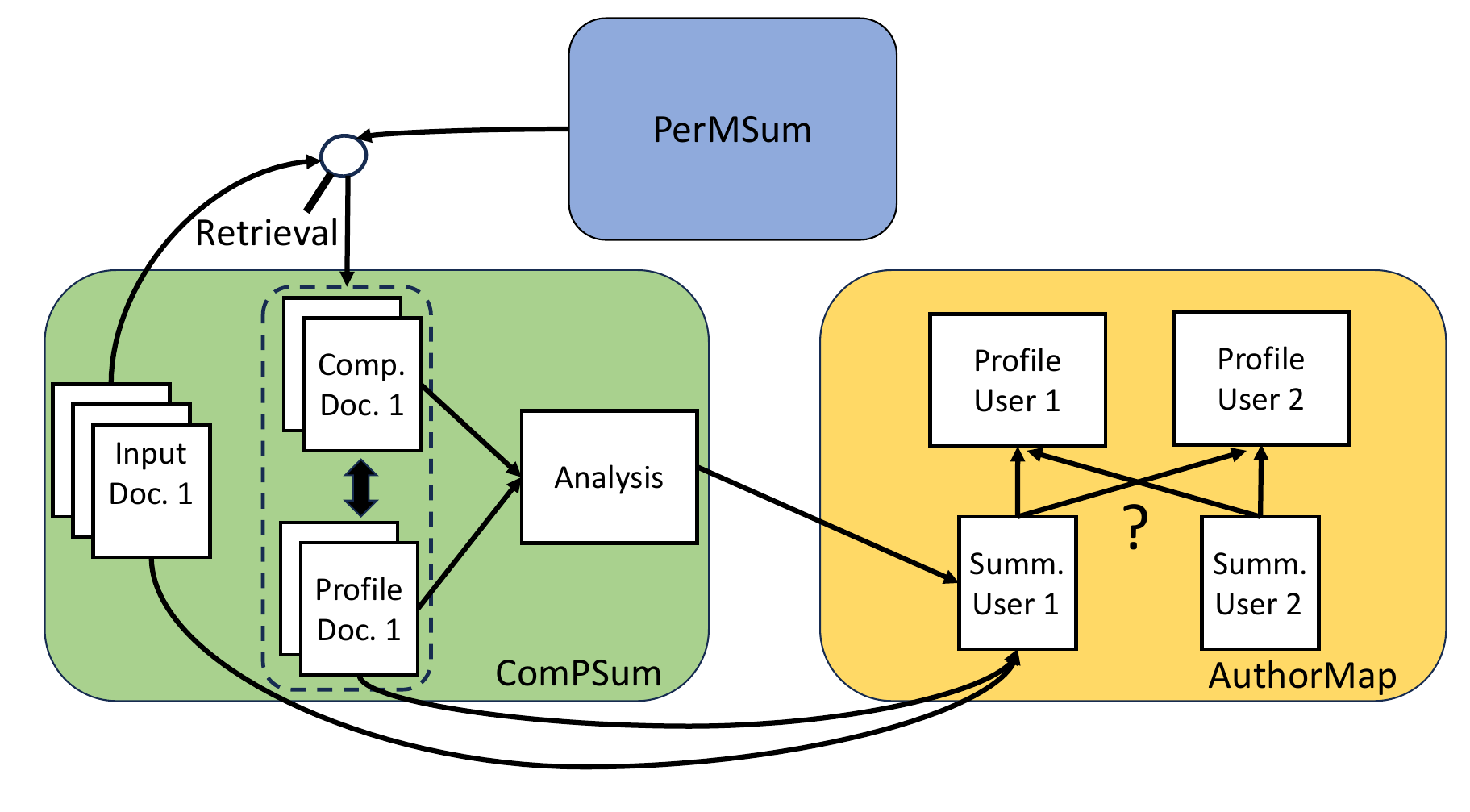}
\caption{Overview of personalized MDS framework, \CPS, and reference-free evaluation framework, \AM.} 
\end{figure}

Motivated by this, we propose \CPS~(\textbf{Com}parative \textbf{P}ersonalization for Multi-Document \textbf{Sum}marization), a personalized MDS framework. 
Specifically, \CPS~considers two key preference dimensions: 
writing style and content focus \cite{zhang2024personalization}. \CPS~first generates a structured analysis of these two dimensions for a user by comparing profile documents with other documents authored by different users on the same topic. \CPS~then uses the generated structured analysis to guide the generation of personalized summaries that capture the user's writing style and content focus. 

Apart from generating personalized summaries, their evaluation is also a major challenge since there is no reference summary and conventional metrics like ROUGE \cite{lin-2004-rouge} cannot be applied. 
To address this issue, we propose \AM, a fine-grained reference-free evaluation framework for personalized MDS that independently assesses writing style and content focus. 
To control the inherent differences between different systems, \AM~evaluates the personalization of a system based on the authorship attribution between two personalized summaries generated for different users by the same system. 
We perform automatic and human evaluation of \AM~and show that it achieves reasonable accuracy. 

For robust evaluation of personalized MDS, we construct \PM, an MDS dataset spanning reviews and news domains. For evaluation, we need documents labeled with their authors (users). 
While there are MDS datasets with user labels in the review domain \cite{ni-etal-2019-justifying}, the news domain still lacks such datasets. To construct such a dataset, we use news articles from the All The News dataset \footnote{https://components.one/datasets/all-the-news-2-news-articles-dataset/}. We collect and process 14K document sets and 1.4K users each of whom authored at least 10 documents. These document sets and corresponding author information are then combined with the document sets sampled from the Amazon dataset \cite{ni-etal-2019-justifying} to form \PM, which results in 45K document sets and 5.3K users. 

Using \AM, 
we find that \CPS~achieves consistent improvement on \PM~with different LLMs while maintaining other critical qualities of summaries, such as relevance and factuality. 

Our contributions are four-fold: 
\begin{itemize}[topsep=1pt, leftmargin=*, noitemsep]
    \itemsep0mm
    \item We propose \CPS, a personalized MDS framework based on comparative personalization; 
    \item We propose \AM, a fine-grained reference-free evaluation framework for personalized MDS;
    \item We propose \PM, a personalized MDS dataset in the news and review domain;
    \item We evaluate the performance of \CPS~on \PM~using \AM, showing that it outperforms strong baselines.
    
\end{itemize}

\section{Related Work}
Personalized text generation aims to generate a personalized text for a given user based on their profile documents. Recent works address personalized text generation using user's profile documents. For example, \citet{salemi2024lamp} retrieve related documents from a user's profile and \citet{li2023teach} train models to summarize and synthesize the retrieved documents. To address the information loss problem in retrieval, \citet{richardson2023integrating} additionally include a summary of user profile. However, these works generally model user individually. Recently, \citet{sun2025persona} use other similar users' profile to infer a user's profile when existing data about the user is sparse. In a recent work, \citet{qiu2025measuring} improve personalized review generation by comparing a user's review with other user's review. However, their design is specific to the review domain and may not generalize well to other domains. We perform experiments to show that \CPS~outperforms their method in Sec. \ref{evaluation}.


Most previous works on personalized text generation \cite{salemi2024lamp, ao2021pens} use reference-based metrics that evaluate the similarity between generated texts and human-written references, such as ROUGE \cite{lin-2004-rouge} and BERTScore \cite{zhang2019bertscore}. Recently, \citet{salemi2025experteffectiveexplainableevaluation}, proposes to evaluate personalized text generation from writing style and content, which are similar to the dimensions used by \AM, but it still needs access to reference texts. For reference-free evaluation, \citet{wang2023automated} \ propose AuPEL, which evaluates personalization based on pairwise authorship attribution \cite{bozkurt2007authorship} between two personalized texts from different systems for the same user. 
However, AuPEL overlooks various dimensions of personalization and does not control the inherent differences in writing styles or content focuses between different systems. \citet{zhang2025rehearse} uses aspect and sentiment similarity between personalized summaries and user profiles to evaluate personalization. However, their design is specific to the review domain and may not generalize well to other domains. 
 
 \section{Problem Statement}
The input of personalized MDS is a document set $D$ containing multiple documents on the same topic to be summarized. For personalization for user $u$, the input also contains a profile $P_u$ containing multiple profile documents $p_u^i \in P_u$ authored by the user $u$. 
Given these inputs, the output of personalized MDS is a personalized summary $s_u$ of the document set $D$ that capture the individual preference of user $u$ as expressed in their profile $P_u$.

\section{\CPS}
In this section, we describe our proposed framework for personalized MDS, \CPS. We first describe how \CPS~generates a structured analysis $a_u$ of a user $u$ that captures their distinctive features of writing styles and contents focuses by comparing with documents written by other users. We then describe how it uses the structured analysis $a_u$ to generate a personalized summary $s_u$.   

\begin{table}[]
\footnotesize
\centering
\begin{tabular}{p{7.2cm}}
\toprule
\textbf{Style analysis}: User X's writing style is characterized by its concise and straightforward narrative, focusing on… \\
\textbf{Content analysis}: User X's profile texts tend to focus on personal reflections and positive developments. Unlike other users,  User X avoids 
sensationalism …  \\
\bottomrule
\end{tabular}
\caption{Structured analysis of a user on dimensions of writing style and content focus.}
\label{tab:structure}
\end{table}

\noindent\textbf{Generating structured analysis:} \CPS~uses an LLM to generate structured analysis, $a_u$, of a user $u$ from two dimensions: writing style and content focus. Tab. \ref{tab:structure} shows an example. 
\CPS~explicitly focuses on these two dimensions so that it only captures preferences but not unrelated information, like a general summary of each profile document. 
To generate the structured analysis, following \citet{salemi2024lamp}, \CPS~first retrieves the top $k$ documents from user $u$'s profiles $P_u$ using a retrieval model $\mathcal{R}$. 
For the retrieval query, \CPS~uses the concatenation of all documents belonging to document set $D$ and retrieves $k$ profile documents most similar to the query: $\mathcal{R}(D,P_u,k)$. 

However, generating the structured analysis $a_u$ only based on the retrieved profile documents can make the analysis focus on general features of the user $u$ but ignore finer differences compared to other users. To address this issue, comparing documents on the same topic written by other users can be useful. Therefore, for each retrieved profile document $p_u^i\in\mathcal{R}(D,P_u,k)$, \CPS~identifies a set of documents, $C^{p_i}_{\neg u}$, that belong to the same document set as $p_u^i$ (and hence are on the same topic) but are written by different users. 
\CPS~then retrieves one comparative document $p_{\neg u}^i \in C^{p_i}_{\neg u}$ that is most dissimilar to $p_i$ using the retrieval model $\mathcal{R}$. 
By comparing every pair of profile document $p_i$ and its comparative document $p_{\neg u}^i$, \CPS~then instructs an $LLM$ to generate the structured analysis $a_u$ that focuses on distinctive features of writing style and content focus that set the user $u$ apart:
\vspace{-0.15cm}
\begin{equation}
a_u=LLM(p_u^1,p_{\neg u}^1,...,p_u^k,p_{\neg u}^k)
\end{equation}
\noindent\textbf{Generating personalized summary:}
Using this structured analysis $a_u$ and the retrieved profile documents $p^1_u,...p^k_u$, \CPS~generates a personalized summary $s_u$ for the document set $D$:
\vspace{-0.15cm}
\begin{equation}
s_u=LLM(p^1_u,...,p^k_u,a_u,D)
\end{equation}
Specifically, \CPS~instructs the summarizing $LLM$ to generate a summary $s_u$ that mimics the writing style and content focus based on retrieved profile documents $p^1_u,...,p^k_u$ and structured analysis $a_u$, while ensuring that the summary $s_u$ includes only contents presented in the document set $D$. We show the prompts used for \CPS~in App. \ref{summ_prompt}. 

\section{\AM}
In this section, we describe our proposed fine-grained reference-free evaluation framework, \AM.
\AM~evaluates personalization along two key dimensions: writing style and content focus. 
The underlying idea behind \AM~ is that if the generated summary is well-personalized, it will be possible to infer the user's preferences from the summary and use them for the task of authorship attribution. 
\AM~considers two profiles, $P_{u_1}$ and $P_{u_2}$, and two personalized summaries, $s_{u_1}$ and $s_{u_2}$, of the same input document set, $D$, generated for user $u_1$ and $u_2$ respectively. \AM~then evaluates whether each profile $P_{u_*}$ can be correctly attributed to its user based on personalized summaries. 

\AM~performs such evaluation separately based on writing style and content focus. However, users can have different preferences for the two dimensions on different topics. Therefore, when evaluating based on a certain dimension, \AM~first retrieves the top $n$ profile documents from each user's profile $P_{u_*}$ most similar to the concatenation $s_{u_1}$ and $s_{u_2}$ using the retrieval model $\mathcal{R}$: $\mathcal{R}(s_{u_1} \circ s_{u_2}, P_{u_*}, n)$, where $\circ$ denotes the concatenation. \AM~uses concatenated summaries instead of just one summary to ensure that no summary has inherent advantages. 
For simplicity, we use $P_{u_1}$ and $P_{u_2}$ to denote the retrieved profiles. For each retrieved profile $P_{u_*}$, \AM~instructs a judge LLM to predict which user, $u_1$ or $u_2$, is more likely to be the author of the retrieved profile $P_{u_*}$ given their personalized summaries, $s_{u_1}$ and $s_{u_2}$ for the given dimension:
\vspace{-0.15cm}
\begin{align}
\hat{u}_1= LLM_{judge}(P_{u_1},s_{u_1},s_{u_2})\\
\hat{u}_2= LLM_{judge}(P_{u_2},s_{u_1},s_{u_2})
\end{align}
where $\hat{u}_*$, the predicted author of the profile $P_{u_*}$, can be $u_1$, $u_2$ or tie. If the summary $s_{u_1}$ is well personalized for user $u_1$, the LLM judge will be able to attribute the profile $P_{u_1}$ to user $u_1$. The same should also apply to user $u_2$. 

To mitigate positional bias \cite{huang2023embrace}, \AM~performs such prediction twice with different orders of $s_{u_1}$ and $s_{u_2}$, which results in four predictions in total. \AM~evaluates the personalization capability of a personalized MDS system as the percentage of samples where the judge LLM correctly predicts the author of the retrieved profile in the majority of four predictions. A larger percentage value indicates better personalization capability on the corresponding dimension of the personalized MDS system. The prompts for \AM~are shown in App. \ref{eval_prompt}.

\section{\PM}
In this section, we describe how we construct the \PM~dataset. We first describe how \PM~obtains document sets where each document in a set is labeled with a user (its author).
We then describe how to select samples from \PM~for the evaluation using \AM.

\noindent\textbf{Obtaining Document Set with User Label:} To obtain document sets with user labels in the news domain, \PM~uses the All the News dataset, which includes details such as author names, publishing media, and publishing dates. However, the news articles from this dataset have two issues for direct application in personalized MDS and its evaluation. First, some news articles contain explicit mentions of the author or media outlet (e.g., “XXX reports in New York”). These direct mentions are undesirable shortcuts for personalized summarization and its evaluation. To address this issue, \PM~removes all sentences containing author names or publishing media. 
Second, some news articles have more than three authors or media organizations as authors. These news articles might not truly reflect the preference of their individual authors. Therefore, \PM~ labels them accordingly so that they are not used as profile documents. 
After labeling documents with users, \PM~clusters documents into document sets based on token overlap, named entities, and publishing dates as \citet{liu2022politics} so that documents in each document set are about the same event. 

To obtain document sets with user labels in the review domain, \PM~uses reviews from the book category of the Amazon dataset \cite{ni-etal-2019-justifying} following \citet{wang2023automated}. \PM~ preprocesses the reviews and obtains document sets following \citet{bravzinskas2019unsupervised} by only keeping reviews that are between $50$ to $150$ words and are written in English. \PM~also filters out reviews written by users who write more than $200$ reviews. 

For both domains, \PM~only considers users that write at least $10$ documents and splits the users into training, validation, and test sets using the user split motivated by \citet{salemi2024lamp}. To prevent information leakage, document sets are also split into training, validation, and test sets. Hence, there is no overlap between users or document sets in the three splits of the dataset. More details of data curation process are in App. \ref{preprocess_dataset}. 

\noindent\textbf{Sample selection for \AM:} Each sample for evaluation with \AM~requires a document set $D$ and two personalized summaries for users $u_1$ and $u_2$. However, randomly sampling two users for evaluation can face the issue of sparsity in personalization \cite{dong-etal-2024-llm}. For example, for a user whose profile documents are all entertainment news, it can be difficult to get enough information from their profile for generating a personalized summary of international news. To address this issue, for each document set $D$, \PM~only selects pairs of users $u_1$ and $u_2$ who write documents belonging to the document set $D$. 
However, in such cases, when generating a personalized summary $s_{u_*}$ for a user $u_*$, the personalized MDS systems might identify and copy information from the input documents written by the user $u_*$. 
To prevent this, we remove all documents written by the users from the input document set. The generated pairs of personalized summaries are then used for evaluation using \AM. To prevent users from dominating the evaluation, \PM~limits each user to appear in at most $100$ samples. Statistics of \PM~are reported in Tab. \ref{tab:stat}. 

\begin{table*}[ht!]
\centering
\small
\resizebox{0.9\textwidth}{!}{
\begin{tabular}{lcccccc}
\toprule
       & \#User       & \#Doc. Set  & \#Sample  & Prof. Size & Doc. Set. Size  & Doc. Len.  \\ \midrule
News   & 828/293/296  & 10730/1393/1463 & -/2085/2360 & 39.72 & 3-10 & 216.72 \\ 
Review & 2400/763/766 & 27725/1878/1795 & -/2774/2757 & 19.77 & 8 & 86.40 \\ \bottomrule
\end{tabular}}
\caption{Statistics of \PM. We report Numbers of users, document sets, and samples in training, validation, and test sets. We also report average profile size per user, size of input document sets, and length of documents. }
\label{tab:stat}
\end{table*}
\section{Experiments}
In this section, we describe experiments on \AM~and \CPS.
\subsection{Implementation Details}
For \AM, we use Llama3.3-70b-Instruct \cite{llama3modelcard} as the LLM judge. \AM~can also use other LLMs as the LLM judge. We perform evaluation using Gemma-3-27b-it \cite{gemma_2025} in App. \ref{app:gemma3_eval} and find that \AM~is independent of the choice of the LLM judge. For authorship attribution, \AM~retrieves $n=5$ profile documents using BM25 \cite{robertson1995okapi}. All retrieved profile documents are truncated to $100$ words to control the impact of document length. Motivated by \citet{huang2024can}, when evaluating writing style, the prompt additionally instruct the LLM judge to focus on lingustic features like modal verbs and typos. 

For \CPS, we experiment with Llama3.1-8b-Instruct \cite{llama3modelcard}, Qwen2.5-14B-Instruct \cite{qwen2}, and Llama3.3-70b-Instruct. For personalization, \CPS~retrieves $m=5$ profile documents and comparative documents using BM25. The token limit for personalized summaries is 100 words. To match the token limit,  all retrieved profile documents are also truncated to $100$ words. All LLMs used in experiments use the default sampling parameter. Hyperparameters and prompts are tuned on the validation set. 

\subsection{Automatic Evaluation of \AM}
In this section, we perform automatic evaluation to evaluate the accuracy of \AM. The straightforward way is to evaluate its accuracy on reference personalized summaries, but they are not available. However, for evaluating \AM, reference summaries are not the only choice. Therefore, for this evaluation, we use documents from the same document set but written by two users of interest. 
To mimic the setting of \AM, the documents are truncated to $100$ words, matching the length of a typical generated personalized summary. We report the accuracy of \AM~on human-written documents in the test sets of \PM. For comparison, we also report the accuracy of \AM~when profile documents $P_{u_*}$  are not retrieved but randomly sampled from user profiles. This setup resembles the setting of \citet{wang2023automated}. The results are shown in Tab. \ref{tab:am_eval}. 

\begin{table}[]
\centering
\small
\resizebox{0.48\textwidth}{!}{
\begin{tabular}{lcccc}
\toprule
                   & \multicolumn{2}{c}{News} & \multicolumn{2}{c}{Review} \\

                   & style      & content     & style       & content      \\ \midrule
\AM & 76.65      & 71.64       & 89.00       & 82.69        \\
~~w/o retrieval      & 75.10      & 68.71       & 87.88       & 81.27 \\
\bottomrule
\end{tabular}}
\caption{Automatic Evaluation of \AM. \AM~shows reasonable accuracy on classifying human written documents and outperforms its variant without retrieval. }
\label{tab:am_eval}
\end{table}

From the table, we observe that \AM~shows reasonable accuracy on the documents, suggesting that it can reliably evaluate personalization in different dimensions. Besides, \AM~outperforms its variant without retrieval, which is used by \citet{wang2023automated}, showing that retrieval is useful to capture the varying preferences of users on different topics.  

 Writing style and content focus are independent dimensions \cite{jafaritazehjani-etal-2020-style} and \AM~should be able to evaluate them independently. Therefore, for a more controlled evaluation, we test the accuracy of \AM~on paraphrased documents with altered writing styles. Specifically, we evaluate the accuracy of \AM~on two types of document pairs : $d_{u_1}$ vs $para(d_{u_2})$, and $d_{u_1}$ vs $para(d_{u_1})$, where $d_{u_*}$ denotes the original document written by user $u_*$, $para(d_{u_*})$ denotes a paraphrased document following the writing style of the other user. The first document pair help us evaluate from the perspective of the content focus with similar writing styles. The second document pair help us evaluate from the perspective of the writing style with similar content focuses. We use LLama3.3-70b-Instruct to paraphrase the human-written documents to mimic the writing style of the other user. The prompt for paraphrasing is shown in App. \ref{para_prompt}. Please note that the paraphrasing may not be perfect as it may not completely mimic the writing style of the given user. However, it is sufficient for us to test the independence based on changes in \AM's accuracies. If \AM~evaluates writing style and content focus independently, \AM~should show higher accuracy on $d_{u_1}$ vs $para(d_{u_2})$ when evaluating content focus since the document pair has differ in content focus but not much in style. Conversely, when evaluating writing style, \AM~should show higher accuracy on $d_{u_1}$ vs $para(d_{u_1})$ since the document pairs differ in style but not much in content focus. The results are shown in Tab. \ref{tab:am_ind}.
\begin{table}[]
\centering
\small
\resizebox{0.48\textwidth}{!}{
\begin{tabular}{lcccc}
\toprule
                                  & \multicolumn{2}{c}{News} & \multicolumn{2}{c}{Review} \\
               & style      & content     & style       & content      \\ \midrule
$d_{u_1}$ vs $para(d_{u_2})$ & 61.34      & 58.76       & 70.34       & 70.58        \\
$d_{u_1}$ vs $para(d_{u_1})$ & 68.77      & 31.68       & 77.93       & 54.10       \\
\bottomrule
\end{tabular}}
\caption{Accuracy of \AM~on paraphrased human written documents. The results show that \AM~can independently evaluates writing style and content focus.}
\label{tab:am_ind}
\end{table}

From the table, we observe the desirable pattern: \AM~shows higher accuracy on $d_{u_1}$ vs $para(d_{u_2})$ when evaluating content focus and higher accuracy on $d_{u_1}$ vs $para(d_{u_1})$ when evaluating writing style. The result shows that \AM~can evaluate personalization based on either writing style or content focus independently.

\subsection{Human Evaluation of \AM}

We perform a human evaluation to further evaluate the correlation between \AM~and human judgment. Following the practice of \AM, each human evaluation sample contain two personalized summaries generated for different users and a set of retrieved profile documents written by one of the user. For each sample, human annotators are instructed to select which summary better aligns with the writing style or content focus expressed by the profile documents. We then calculate the accuracy of \AM~ by measuring proportion of samples where \AM~and human annotation select the same summary.

We randomly sample $15$ samples from each domain, news and reviews, yielding $30$ samples in total. Personalized summaries of each sample are generated by \CPS~with Llama3.1-8b, which shows the medium-level performance in Tab. \ref{tab:cps_overall}. Each sample is annotated by three annotators recruited from Amazon Mechanical Turk. More details of human evaluation are in App. \ref{app:human_eval}. 

Among $30$ samples, the Randolph’s Kappa \cite{randolph2005free} between three annotators is $0.40$, showing a moderate correlation. On the news domain, \AM~achieves $80$ percent accuracy for writing styles and $73$ percent for content focus. On the review domain, \AM~achieves $73$ percent accuracy for writing style and $80$ percent for content focus. The results show that \AM~achieves high accuracy when compared to human annotated labels. 


\subsection{Evaluation of \CPS}
\label{evaluation}
\begin{table*}[t]
\centering
\small
\resizebox{0.95\textwidth}{!}{
\begin{tabular}{lcccccccccc}
\hline
\textbf{}   & \multicolumn{5}{c:}{News}                                      & \multicolumn{5}{c}{Review}                \\
            & style & content & fact. & rele. & \multicolumn{1}{c:}{overall} & style & content & fact. & rele. & overall \\ \hline
            & \multicolumn{10}{c}{\textit{Llama3.1-8b-Instruct}}                                                                             \\
RAG         & 53.35                     & 49.61                       & 98.00                          & 96.05                         & \multicolumn{1}{c:}{70.65}                       & 52.52                & 54.12                      & 98.16                       & 92.71                          & 71.32                         \\
CICL        & 53.22                     & 48.89                       & 97.34                          & 95.47                         & \multicolumn{1}{c:}{70.12}                       & 54.56                & 55.83                      & 96.87                       & 88.91                          & 71.57                         \\
RAG+Summary & 54.58                     & 50.89                       & 98.03                          & 97.17                         & \multicolumn{1}{c:}{71.72}                       & 54.56                & 57.10                      & 97.67                       & 92.02                          & 72.74                         \\
DPL         & 53.94                     & 47.90                       & 97.91                          & 96.30                         & \multicolumn{1}{c:}{70.26}                       & 54.70                & 59.13                      & 97.05                       & 87.68                          & 72.43                         \\
Rehersal    & 99.36                     & 99.49                       & 23.16                          & 28.75                         & \multicolumn{1}{c:}{50.65}                       & 97.97                & 98.40                      & 57.28                       & 37.62                          & 67.51                         \\
CompSum     & 59.75                     & 53.94                       & 98.01                          & 95.32                         & \multicolumn{1}{c:}{\textbf{74.07}}                       & 59.13                & 57.89                      & 98.03                       & 91.99                          & \textbf{74.54}  \\ \hdashline
            & \multicolumn{10}{c:}{\textit{Qwen2.5-14B-Instruct}}                                                                                  \\
RAG         & 52.88                     & 48.26                       & 98.07                          & 96.77                         & \multicolumn{1}{c:}{70.15}                       & 48.78                & 51.18                      & 97.76                       & 91.76                          & 68.79                         \\
CICL        & 51.69                     & 48.64                       & 97.11                          & 96.68                         & \multicolumn{1}{c:}{69.71}                       & 54.66                 & 55.90                      & 96.95                       & 88.91                          & 71.64                          \\
RAG+Summary & 52.75                     & 49.32                       & 98.17                          & 97.93                         & \multicolumn{1}{c:}{70.72}                       & 54.88                & 59.75                      & 97.40                       & 90.97                          & 73.42                         \\
DPL         & 50.97                     & 46.31                       & 98.04                          & 97.37                         & \multicolumn{1}{c:}{68.90}                       & 50.71                & 57.53                      & 96.83                       & 89.42                          & 70.89                         \\
Rehersal    & 98.69                     & 99.41                       & 22.86                          & 26.94                         & \multicolumn{1}{c:}{49.58}                       & 95.79                & 98.69                      & 53.20                       & 33.63                          & 64.13                         \\
CompSum     & 57.92                     & 57.08                       & 97.96                          & 96.58                         & \multicolumn{1}{c:}{\textbf{74.78}}                       & 60.36                & 63.92                      & 96.51                       & 89.98                          & \textbf{76.08}                         \\ 

\hdashline
            & \multicolumn{10}{c:}{\textit{Llama3.3-70b-Instruct}}                                                                                  \\
RAG         & 48.64                     & 39.96                       & 98.59                          & 97.98                         & \multicolumn{1}{c:}{65.83}                       & 50.82                & 46.13                      & 98.76                       & 93.16                          & 68.15                         \\
CICL        & 50.04                     & 44.41                       & 98.55                          & 98.03                         & \multicolumn{1}{c:}{68.07}                       & 49.36                & 52.09                      & 98.70                       & 92.79                          & 69.66                         \\
RAG+Summary & 49.96                     & 41.57                       & 98.70                          & 98.08                         & \multicolumn{1}{c:}{66.96}                       & 52.99                & 47.80                      & 98.68                       & 93.43                          & 69.52                         \\
DPL         & 49.32                     & 41.40                       & 98.81                          & 98.17                         & \multicolumn{1}{c:}{66.71}                       & 50.96                & 46.21                      & 98.50                       & 93.10                          & 68.17                         \\
Rehersal    & 99.49                     & 100.00                      & 21.74                          & 24.90                         & \multicolumn{1}{c:}{48.17}                       & 97.53                & 98.55                      & 62.03                       & 33.81                          & 67.00                         \\
CompSum     & 53.90                     & 45.38                       & 98.64                          & 98.03                         & \multicolumn{1}{c:}{\textbf{69.74}}                       & 59.27                & 51.87                      & 98.70                       & 93.35                          & \multicolumn{1}{c:}{\textbf{72.95}}                         \\ \hline
\end{tabular}}
\caption{Evaluation of \CPS. A higher value indicates better performance. The best-performing method based on overall score is \textbf{bolded}. \CPS~shows the best overall performance. }
\label{tab:cps_overall}
\end{table*}
In this section, we evaluate the qualities of personalized summaries generated by \CPS. To evaluate this, we consider both personalization as well as general qualities of summaries. For personalization, we consider two dimensions: writing style and content focus using \AM. For general qualities, we consider factuality, which measures whether summaries only contain information supported by the input document set, and relevance, which measures whether summaries only include important information from document sets \cite{fabbri2021summeval}. To evaluate factuality, we use FactScore \cite{factscore}. To evaluate relevance, we use G-Eval \cite{liu2023g}. 
To match the scales of other measures, we map its score to 1-100. We additionally report an overall score, which is the arithmetic average of the four measures. A higher value indicates better overall quality. 

We compare \CPS~with the following baselines: (i) RAG \cite{salemi2024lamp} that generates a personalized summary based on retrieved profile documents; (ii) CICL \cite{gao2024customizing}, which extends RAG by additionally retrieving comparative documents authored by other users; (iii) RAG+Summ  \cite{li2023teach}, which generates summaries of the retrieved profile documents and uses them to guide generation of personalized summaries; (iv) DPL \cite{qiu2025measuring}, which first generates an analysis for each retrieved profile document by comparing its comparative documents and then aggregate these analyses into a profile summary to generate the personalized summary; (v) Rehearsal, which \cite{zhang2025rehearse} iteratively refines a general summary using a user agent and a supervisor agent. More details for implementation of these baselines are shown in App. \ref{exp_cps_eval}. We report the results of \CPS~and these baselines on the test set of \PM~in Tab. \ref{tab:cps_overall}.

From the table, we observe that \CPS~generally outperforms all baselines on personalization and general summary qualities and also achieves the best overall scores. All differences between \CPS~and the second-best performing methods are statistically significant using paired bootstrap resampling ($p<0.05$) \cite{koehn-2004-statistical}.

Comparing with Rehearsal, we observe that although Rehearsal achieves high performance on personalization, its performance on general qualities is pretty low. We find that many summaries generated by Rehearsal resemble user profiles rather than faithful summaries of input document sets. Further analysis of the `modification suggestions' generated by Rehearsal shows that they often suggest adding information that only exists in the user profile but not input document sets. This can be caused by the fact that neither the user agent nor the supervisor agent has access to input document sets when generating the suggestions. The findings also shows the importance of both personalization and general qualities during evaluation. 

Comparing with DPL, we observe that while both methods use comparative documents when generating analysis of users, \CPS~outperforms DPL especially in the news domain. This can be caused by two reasons. First, DPL is designed specifically for reviews and instructs the LLM to focus on aspects like emotional style that do not generalize to domains beyond reviews (and in this sense this comparison is not fair to DPL). Second, DPL’s analysis is initially generated from a single profile document, which is generally not enough to infer the preference of a user. Even though DPL later summarizes multiple analyses, information loss is still inevitable. Contrarily, \CPS~generates the structured analysis of a user directly conditioned on multiple profile documents. A fairer comparison with DPL using the same aspect as \CPS~is provided in Sec. \ref{sec:ablation}. 

\subsection{Ablation Study of \CPS}
\label{sec:ablation}
\begin{table}[]
\centering
\resizebox{0.48\textwidth}{!}{
\begin{tabular}{lccccc}
\toprule
                    & \multicolumn{2}{c}{Llama3.1-8b} & \multicolumn{2}{c}{Qwen2.5-14b} &                \\
                    & News           & Review         & News           & Review         & Avg.        \\ \midrule
\CPS             & \textbf{74.08} & 74.54          & \textbf{74.78} & 76.08          & \textbf{74.87} \\
~~w/o comp. doc.  & 71.10          & 73.24          & 72.34          & 76.10          & 73.19          \\
~~w/o structure   & 71.43          & \textbf{78.43} & 69.34          & 75.88          & 73.77          \\
~~w/ sim. comp.   & 73.33          & 74.00          & 71.84          & \textbf{76.85} & 74.00          \\
~~w/ multi. stage & 69.87          & 75.15          & 68.62          & 74.43          & 72.02          \\
\bottomrule
\end{tabular}}
\caption{Overall performance of \CPS~and its ablated variants. The best-performing method is \textbf{bolded}. \CPS~outperforms its ablated version, showing the effectiveness of \CPS~design.}
\label{tab:ablation}
\end{table}
In this section, we validate the design of \CPS~by comparing it with the following ablated variants: (i) \textbf{w/o comp. doc.}, which generates structured analysis of a user without comparative documents; (ii) \textbf{w/o structure}, which instructs LLM to generate a profile summary of a user instead of separate analysis of writing style and content focus; (iii) \textbf{w/ sim. comp.}, which generates the structured analysis based on comparative documents that are most similar to profile documents; (iv) \textbf{w/ multi. stage}, which generates the structured analysis in multiple stages, similar to DPL, but focusing on writing style and content focus instead of dimensions used by DPL. For these ablated variants, we report their overall scores in Tab. \ref{tab:ablation} which evaluates personalization and general qualities as described in Sec. \ref{evaluation} on the test set of \PM. The detailed scores for each dimension and implementation details are shown in App. \ref{exp_cps_eval}. 

From the table, we observe that \CPS~outperforms w/o comp. doc. and w/o structure, which shows the effectiveness of comparative documents and structure constraints. Besides, \CPS~also outperforms w/ multi. stage, which shows that directly generating analysis based on multiple profile documents is more effective than generating multiple analyses based on one profile document and summarizing them afterward. Overall, \CPS~outperforms all of its ablated version based on average performance across LLMs and datasets. We show examples of summaries generated by \CPS~in App. \ref{sumamry_example}. We also perform ablation study on number of retrieved documents for \CPS~in App. \ref{app:number_retrieval} and find that retrieve $5$ profile documents performs the best.  

\subsection{Analysis Generated by \CPS}
\label{analysis}
\begin{table}[]
\centering
\resizebox{0.48\textwidth}{!}{
\begin{tabular}{lccccc}
\toprule
                    & \multicolumn{2}{c}{Llama3.1-8b} & \multicolumn{2}{c}{Qwen2.5-14b} &                \\
                    & News           & Review         & News           & Review         & Avg.        \\ \midrule
\CPS & 80.93 & 82.63 & 78.81 &  81.18 & 80.89 \\
~~w/o comp. doc.      & 83.18          & 82.74          & 81.73          & 82.42          & 82.52       \\
\bottomrule
\end{tabular}}
\caption{Average similarity between structured analysis of different users. Comparative documents can make structured analysis more diverse for different users.}
\label{tab:cos_sim}
\end{table}
In this section, we examine whether using comparative documents leads to more diverse structured analysis for different users. To evaluate this, for each sample in the test set, we measure the average similarity between the structured analysis generated for two users for the same input document set. We then compare the similarity scores produced by \CPS~with those from its ablated variant, \textbf{w/o comp. doc.}, which generates structured analysis without using comparative documents. To measure the similarity, we report cosine similarity of structured analysis's embedding generated by gte-Qwen2-1.5B-instruct \cite{li2023towards} in Tab. \ref{tab:cos_sim}. 

From the table, we observe that \CPS~has lower similarity score than w/o comp. doc., suggesting that using comparative documents leads to more diverse structured analysis for different users. 
We also show examples of structured analysis for writing style generated by \CPS~and w/o comp. doc. in Tab. \ref{tab:example}. From the examples, we can observe that the structured analysis generated by \CPS~additionally includes comparison with other users, which helps the MDS system to better differentiate different users. We show additional examples of structured analysis in App. \ref{qualitative_example}.

\begin{table}[]
\footnotesize
\centering
\begin{tabular}{p{7.4cm}}
\toprule
\textbf{\CPS}: User X's writing style is characterized by a clear and concise narrative voice, often incorporating direct quotes and specific details... \textbf{Unlike other users, who may rely on sensational language or emotional appeals}, User X's tone is measured and informative, making their content feel more authoritative and trustworthy. \\ \midrule
\textbf{w/o comp. doc.}: User X's writing style is characterized by a conversational tone and a focus on storytelling. They often use anecdotes and quotes from celebrities to illustrate their points, making their content feel more relatable and engaging. The text is also well-structured and easy to follow, with a clear and concise writing style...\\
\bottomrule
\end{tabular}
\caption{Structured analysis for writing style generated by \CPS~and w/o comp. doc. The structured analysis generated by \CPS~includes comparison with other users (in \textbf{bold}), which helps in better personalization.}
\label{tab:example}
\end{table}
\section{Conclusion}
We propose \CPS, a personalized MDS framework. It captures the finer differences between users by comparing profile documents with other documents authored by different users. We also propose \AM, a reference-free fine-grained evaluation framework. We perform automatic and human evaluation to evaluate \AM~and show that it achieves reasonable accuracy. For robust evaluations of \CPS, we construct \PM, a personalized MDS dataset in the news and review domain. We evaluate the performance of \CPS~on \PM~using \AM, showing that it outperforms strong baselines.
\section{Limitation}
One limitation of \CPS~is its reliance on comparable documents that share the same topic as the profile documents. Otherwise, the differences between profile documents and comparable can stem from topic difference but not individual preference differences. Therefore, \CPS~cannot be directly applied to general personalized text generation as identifying such comparable documents can be difficult for other tasks. Future work could explore methods for automatically identifying or generating comparable documents to broaden the applicability of \CPS~to more general personalized text generation

When constructing the \PM~dataset, we define the profile documents for a user as documents written by the user. However, for the news domain, defining the profile documents as documents clicked or liked by the user seems to be more similar to the real-world application. Unfortunately, existing publicly available news datasets with user interaction data are not suitable for personalized MDS. For example, the PENS dataset \cite{ao2021pens} lacks publishing dates of news articles, which makes it difficult to efficiently create input document sets where all documents are about the same event. MIND dataset \cite{wu2020mind} lacks both publishing dates and full texts of news articles. Therefore, we construct \PM~using articles from All the News dataset, which is the only large-scale dataset that includes publishing dates, author information, and full article texts.
\section{Ethical Consideration}
The datasets we use are all publicly available.  All the models used in this paper are publicly accessible. The inference and finetuning of models are performed on four Nvidia A6000 or Nvidia L40 GPUs. We do not annotate any data on our own.

We perform human evaluation experiments on Amazon Mechanical Turk. The annotators were compensated at a rate of \$20 per hour. During the evaluation, human annotators were not exposed to any sensitive or explicit content.
\bibliography{acl_latex}

\begin{thebibliography}{36}
\providecommand{\natexlab}[1]{#1}

\bibitem[{AI@Meta(2024)}]{llama3modelcard}
AI@Meta. 2024.
\newblock \href {https://github.com/meta-llama/llama3/blob/main/MODEL_CARD.md} {Llama 3 model card}.

\bibitem[{Ao et~al.(2021)Ao, Wang, Luo, Qiao, He, and Xie}]{ao2021pens}
Xiang Ao, Xiting Wang, Ling Luo, Ying Qiao, Qing He, and Xing Xie. 2021.
\newblock Pens: A dataset and generic framework for personalized news headline generation.
\newblock In \emph{Proceedings of the 59th Annual Meeting of the Association for Computational Linguistics and the 11th International Joint Conference on Natural Language Processing (Volume 1: Long Papers)}, pages 82--92.

\bibitem[{Bozkurt et~al.(2007)Bozkurt, Baghoglu, and Uyar}]{bozkurt2007authorship}
Ilker~Nadi Bozkurt, Ozgur Baghoglu, and Erkan Uyar. 2007.
\newblock Authorship attribution.
\newblock In \emph{2007 22nd international symposium on computer and information sciences}, pages 1--5. IEEE.

\bibitem[{Bra{\v{z}}inskas et~al.(2019)Bra{\v{z}}inskas, Lapata, and Titov}]{bravzinskas2019unsupervised}
Arthur Bra{\v{z}}inskas, Mirella Lapata, and Ivan Titov. 2019.
\newblock Unsupervised opinion summarization as copycat-review generation.
\newblock \emph{arXiv preprint arXiv:1911.02247}.

\bibitem[{Bra{\v{z}}inskas et~al.(2020)Bra{\v{z}}inskas, Lapata, and Titov}]{bravzinskas2020unsupervised}
Arthur Bra{\v{z}}inskas, Mirella Lapata, and Ivan Titov. 2020.
\newblock Unsupervised opinion summarization as copycat-review generation.
\newblock In \emph{Proceedings of the 58th Annual Meeting of the Association for Computational Linguistics}, pages 5151--5169.

\bibitem[{Dong et~al.(2024)Dong, Hu, and Collier}]{dong-etal-2024-llm}
Yijiang~River Dong, Tiancheng Hu, and Nigel Collier. 2024.
\newblock \href {https://doi.org/10.18653/v1/2024.findings-emnlp.592} {Can {LLM} be a personalized judge?}
\newblock In \emph{Findings of the Association for Computational Linguistics: EMNLP 2024}, pages 10126--10141, Miami, Florida, USA. Association for Computational Linguistics.

\bibitem[{Fabbri et~al.(2021)Fabbri, Kry{\'s}ci{\'n}ski, McCann, Xiong, Socher, and Radev}]{fabbri2021summeval}
Alexander~R Fabbri, Wojciech Kry{\'s}ci{\'n}ski, Bryan McCann, Caiming Xiong, Richard Socher, and Dragomir Radev. 2021.
\newblock Summeval: Re-evaluating summarization evaluation.
\newblock \emph{Transactions of the Association for Computational Linguistics}, 9:391--409.

\bibitem[{Fabbri et~al.(2019)Fabbri, Li, She, Li, and Radev}]{fabbri2019multi}
Alexander~Richard Fabbri, Irene Li, Tianwei She, Suyi Li, and Dragomir Radev. 2019.
\newblock Multi-news: A large-scale multi-document summarization dataset and abstractive hierarchical model.
\newblock In \emph{Proceedings of the 57th Annual Meeting of the Association for Computational Linguistics}, pages 1074--1084.

\bibitem[{Gao and Das(2024)}]{gao2024customizing}
Xiang Gao and Kamalika Das. 2024.
\newblock Customizing language model responses with contrastive in-context learning.
\newblock In \emph{Proceedings of the AAAI Conference on Artificial Intelligence}, volume~38, pages 18039--18046.

\bibitem[{Huang et~al.(2024)Huang, Chen, and Shu}]{huang2024can}
Baixiang Huang, Canyu Chen, and Kai Shu. 2024.
\newblock Can large language models identify authorship?
\newblock In \emph{Findings of the Association for Computational Linguistics: EMNLP 2024}, pages 445--460.

\bibitem[{Huang et~al.(2023)Huang, Laban, Fabbri, Choubey, Joty, Xiong, and Wu}]{huang2023embrace}
Kung-Hsiang Huang, Philippe Laban, Alexander~R Fabbri, Prafulla~Kumar Choubey, Shafiq Joty, Caiming Xiong, and Chien-Sheng Wu. 2023.
\newblock Embrace divergence for richer insights: A multi-document summarization benchmark and a case study on summarizing diverse information from news articles.
\newblock \emph{arXiv preprint arXiv:2309.09369}.

\bibitem[{Jafaritazehjani et~al.(2020)Jafaritazehjani, Lecorv{\'e}, Lolive, and Kelleher}]{jafaritazehjani-etal-2020-style}
Somayeh Jafaritazehjani, Gw{\'e}nol{\'e} Lecorv{\'e}, Damien Lolive, and John Kelleher. 2020.
\newblock \href {https://doi.org/10.18653/v1/2020.coling-main.197} {Style versus content: A distinction without a (learnable) difference?}
\newblock In \emph{Proceedings of the 28th International Conference on Computational Linguistics}, pages 2169--2180, Barcelona, Spain (Online). International Committee on Computational Linguistics.

\bibitem[{Jang et~al.(2023)Jang, Kim, Lin, Wang, Hessel, Zettlemoyer, Hajishirzi, Choi, and Ammanabrolu}]{jangpersonalized}
Joel Jang, Seungone Kim, Bill~Yuchen Lin, Yizhong Wang, Jack Hessel, Luke Zettlemoyer, Hannaneh Hajishirzi, Yejin Choi, and Prithviraj Ammanabrolu. 2023.
\newblock Personalized soups: Personalized large language model alignment via post-hoc parameter merging.
\newblock \emph{arXiv preprint arXiv:2310.11564}.

\bibitem[{Koehn(2004)}]{koehn-2004-statistical}
Philipp Koehn. 2004.
\newblock \href {https://aclanthology.org/W04-3250} {Statistical significance tests for machine translation evaluation}.
\newblock In \emph{Proceedings of the 2004 Conference on Empirical Methods in Natural Language Processing}, pages 388--395, Barcelona, Spain. Association for Computational Linguistics.

\bibitem[{Li et~al.(2023{\natexlab{a}})Li, Zhang, Mei, Wang, Hombaiah, Liang, and Bendersky}]{li2023teach}
Cheng Li, Mingyang Zhang, Qiaozhu Mei, Yaqing Wang, Spurthi~Amba Hombaiah, Yi~Liang, and Michael Bendersky. 2023{\natexlab{a}}.
\newblock Teach llms to personalize--an approach inspired by writing education.
\newblock \emph{arXiv preprint arXiv:2308.07968}.

\bibitem[{Li et~al.(2023{\natexlab{b}})Li, Zhang, Zhang, Long, Xie, and Zhang}]{li2023towards}
Zehan Li, Xin Zhang, Yanzhao Zhang, Dingkun Long, Pengjun Xie, and Meishan Zhang. 2023{\natexlab{b}}.
\newblock Towards general text embeddings with multi-stage contrastive learning.
\newblock \emph{arXiv preprint arXiv:2308.03281}.

\bibitem[{Lin(2004)}]{lin-2004-rouge}
Chin-Yew Lin. 2004.
\newblock \href {https://aclanthology.org/W04-1013} {{ROUGE}: A package for automatic evaluation of summaries}.
\newblock In \emph{Text Summarization Branches Out}, pages 74--81, Barcelona, Spain. Association for Computational Linguistics.

\bibitem[{Liu et~al.(2022)Liu, Zhang, Wegsman, Beauchamp, and Wang}]{liu2022politics}
Y~Liu, X~Zhang, D~Wegsman, N~Beauchamp, and L~Wang. 2022.
\newblock Politics: Pretraining with same-story article comparison for ideology prediction and stance detection.
\newblock \emph{Findings of the Association for Computational Linguistics: NAACL 2022}.

\bibitem[{Liu et~al.(2023)Liu, Iter, Xu, Wang, Xu, and Zhu}]{liu2023g}
Yang Liu, Dan Iter, Yichong Xu, Shuohang Wang, Ruochen Xu, and Chenguang Zhu. 2023.
\newblock G-eval: Nlg evaluation using gpt-4 with better human alignment.
\newblock In \emph{Proceedings of the 2023 Conference on Empirical Methods in Natural Language Processing}, pages 2511--2522.

\bibitem[{Min et~al.(2023)Min, Krishna, Lyu, Lewis, Yih, Koh, Iyyer, Zettlemoyer, and Hajishirzi}]{factscore}
Sewon Min, Kalpesh Krishna, Xinxi Lyu, Mike Lewis, Wen-tau Yih, Pang~Wei Koh, Mohit Iyyer, Luke Zettlemoyer, and Hannaneh Hajishirzi. 2023.
\newblock \href {https://arxiv.org/abs/2305.14251} {{FActScore}: Fine-grained atomic evaluation of factual precision in long form text generation}.
\newblock In \emph{EMNLP}.

\bibitem[{Ni et~al.(2019)Ni, Li, and McAuley}]{ni-etal-2019-justifying}
Jianmo Ni, Jiacheng Li, and Julian McAuley. 2019.
\newblock \href {https://doi.org/10.18653/v1/D19-1018} {Justifying recommendations using distantly-labeled reviews and fine-grained aspects}.
\newblock In \emph{Proceedings of the 2019 Conference on Empirical Methods in Natural Language Processing and the 9th International Joint Conference on Natural Language Processing (EMNLP-IJCNLP)}, pages 188--197, Hong Kong, China. Association for Computational Linguistics.

\bibitem[{Qiu et~al.(2025)Qiu, Zhao, Zhang, Bai, Wang, Cheng, Feng, and Chua}]{qiu2025measuring}
Yilun Qiu, Xiaoyan Zhao, Yang Zhang, Yimeng Bai, Wenjie Wang, Hong Cheng, Fuli Feng, and Tat-Seng Chua. 2025.
\newblock Measuring what makes you unique: Difference-aware user modeling for enhancing llm personalization.
\newblock \emph{arXiv preprint arXiv:2503.02450}.

\bibitem[{Randolph(2005)}]{randolph2005free}
Justus~J Randolph. 2005.
\newblock Free-marginal multirater kappa (multirater k [free]): An alternative to fleiss' fixed-marginal multirater kappa.
\newblock \emph{Online submission}.

\bibitem[{Richardson et~al.(2023)Richardson, Zhang, Gillespie, Kar, Singh, Raeesy, Khan, and Sethy}]{richardson2023integrating}
Chris Richardson, Yao Zhang, Kellen Gillespie, Sudipta Kar, Arshdeep Singh, Zeynab Raeesy, Omar~Zia Khan, and Abhinav Sethy. 2023.
\newblock Integrating summarization and retrieval for enhanced personalization via large language models.
\newblock \emph{arXiv preprint arXiv:2310.20081}.

\bibitem[{Robertson et~al.(1995)Robertson, Walker, Jones, Hancock-Beaulieu, Gatford et~al.}]{robertson1995okapi}
Stephen~E Robertson, Steve Walker, Susan Jones, Micheline~M Hancock-Beaulieu, Mike Gatford, et~al. 1995.
\newblock Okapi at trec-3.
\newblock \emph{Nist Special Publication Sp}, 109:109.

\bibitem[{Salemi et~al.(2025)Salemi, Killingback, and Zamani}]{salemi2025experteffectiveexplainableevaluation}
Alireza Salemi, Julian Killingback, and Hamed Zamani. 2025.
\newblock \href {https://arxiv.org/abs/2501.14956} {Expert: Effective and explainable evaluation of personalized long-form text generation}.
\newblock \emph{Preprint}, arXiv:2501.14956.

\bibitem[{Salemi et~al.(2024)Salemi, Mysore, Bendersky, and Zamani}]{salemi2024lamp}
Alireza Salemi, Sheshera Mysore, Michael Bendersky, and Hamed Zamani. 2024.
\newblock Lamp: When large language models meet personalization.
\newblock In \emph{Proceedings of the 62nd Annual Meeting of the Association for Computational Linguistics (Volume 1: Long Papers)}, pages 7370--7392.

\bibitem[{Sun et~al.(2025)Sun, Yang, Reddy, Fung, Chan, Small, Zhai, and Ji}]{sun2025persona}
Chenkai Sun, Ke~Yang, Revanth~Gangi Reddy, Yi~Fung, Hou~Pong Chan, Kevin Small, ChengXiang Zhai, and Heng Ji. 2025.
\newblock Persona-db: Efficient large language model personalization for response prediction with collaborative data refinement.
\newblock In \emph{Proceedings of the 31st International Conference on Computational Linguistics}, pages 281--296.

\bibitem[{Tan et~al.(2024)Tan, Zeng, Tian, Liu, Yin, and Jiang}]{tan-etal-2024-democratizing}
Zhaoxuan Tan, Qingkai Zeng, Yijun Tian, Zheyuan Liu, Bing Yin, and Meng Jiang. 2024.
\newblock \href {https://doi.org/10.18653/v1/2024.emnlp-main.372} {Democratizing large language models via personalized parameter-efficient fine-tuning}.
\newblock In \emph{Proceedings of the 2024 Conference on Empirical Methods in Natural Language Processing}, pages 6476--6491, Miami, Florida, USA. Association for Computational Linguistics.

\bibitem[{Team(2025)}]{gemma_2025}
Gemma Team. 2025.
\newblock \href {https://goo.gle/Gemma3Report} {Gemma 3}.

\bibitem[{Wang et~al.(2023)Wang, Jiang, Zhang, Li, Liang, Mei, and Bendersky}]{wang2023automated}
Yaqing Wang, Jiepu Jiang, Mingyang Zhang, Cheng Li, Yi~Liang, Qiaozhu Mei, and Michael Bendersky. 2023.
\newblock Automated evaluation of personalized text generation using large language models.
\newblock \emph{arXiv preprint arXiv:2310.11593}.

\bibitem[{Wu et~al.(2020)Wu, Qiao, Chen, Wu, Qi, Lian, Liu, Xie, Gao, Wu et~al.}]{wu2020mind}
Fangzhao Wu, Ying Qiao, Jiun-Hung Chen, Chuhan Wu, Tao Qi, Jianxun Lian, Danyang Liu, Xing Xie, Jianfeng Gao, Winnie Wu, et~al. 2020.
\newblock Mind: A large-scale dataset for news recommendation.
\newblock In \emph{Proceedings of the 58th annual meeting of the association for computational linguistics}, pages 3597--3606.

\bibitem[{Yang et~al.(2024)Yang, Yang, Hui, Zheng, Yu, Zhou, Li, Li, Liu, Huang, Dong, Wei, Lin, Tang, Wang, Yang, Tu, Zhang, Ma, Xu, Zhou, Bai, He, Lin, Dang, Lu, Chen, Yang, Li, Xue, Ni, Zhang, Wang, Peng, Men, Gao, Lin, Wang, Bai, Tan, Zhu, Li, Liu, Ge, Deng, Zhou, Ren, Zhang, Wei, Ren, Fan, Yao, Zhang, Wan, Chu, Liu, Cui, Zhang, and Fan}]{qwen2}
An~Yang, Baosong Yang, Binyuan Hui, Bo~Zheng, Bowen Yu, Chang Zhou, Chengpeng Li, Chengyuan Li, Dayiheng Liu, Fei Huang, Guanting Dong, Haoran Wei, Huan Lin, Jialong Tang, Jialin Wang, Jian Yang, Jianhong Tu, Jianwei Zhang, Jianxin Ma, Jin Xu, Jingren Zhou, Jinze Bai, Jinzheng He, Junyang Lin, Kai Dang, Keming Lu, Keqin Chen, Kexin Yang, Mei Li, Mingfeng Xue, Na~Ni, Pei Zhang, Peng Wang, Ru~Peng, Rui Men, Ruize Gao, Runji Lin, Shijie Wang, Shuai Bai, Sinan Tan, Tianhang Zhu, Tianhao Li, Tianyu Liu, Wenbin Ge, Xiaodong Deng, Xiaohuan Zhou, Xingzhang Ren, Xinyu Zhang, Xipin Wei, Xuancheng Ren, Yang Fan, Yang Yao, Yichang Zhang, Yu~Wan, Yunfei Chu, Yuqiong Liu, Zeyu Cui, Zhenru Zhang, and Zhihao Fan. 2024.
\newblock Qwen2 technical report.
\newblock \emph{arXiv preprint arXiv:2407.10671}.

\bibitem[{Zhang et~al.(2019)Zhang, Kishore, Wu, Weinberger, and Artzi}]{zhang2019bertscore}
Tianyi Zhang, Varsha Kishore, Felix Wu, Kilian~Q Weinberger, and Yoav Artzi. 2019.
\newblock Bertscore: Evaluating text generation with bert.
\newblock In \emph{International Conference on Learning Representations}.

\bibitem[{Zhang et~al.(2025)Zhang, He, and Zhou}]{zhang2025rehearse}
Yanyue Zhang, Yulan He, and Deyu Zhou. 2025.
\newblock Rehearse with user: Personalized opinion summarization via role-playing based on large language models.
\newblock \emph{arXiv preprint arXiv:2503.00449}.

\bibitem[{Zhang et~al.(2024)Zhang, Rossi, Kveton, Shao, Yang, Zamani, Dernoncourt, Barrow, Yu, Kim et~al.}]{zhang2024personalization}
Zhehao Zhang, Ryan~A Rossi, Branislav Kveton, Yijia Shao, Diyi Yang, Hamed Zamani, Franck Dernoncourt, Joe Barrow, Tong Yu, Sungchul Kim, et~al. 2024.
\newblock Personalization of large language models: A survey.
\newblock \emph{arXiv preprint arXiv:2411.00027}.

\end{thebibliography}

\appendix

\section{Appendix}
\subsection{Prompt for \CPS}
\label{summ_prompt}
\begin{figure*}
\centering
\includegraphics[width=0.95\textwidth]{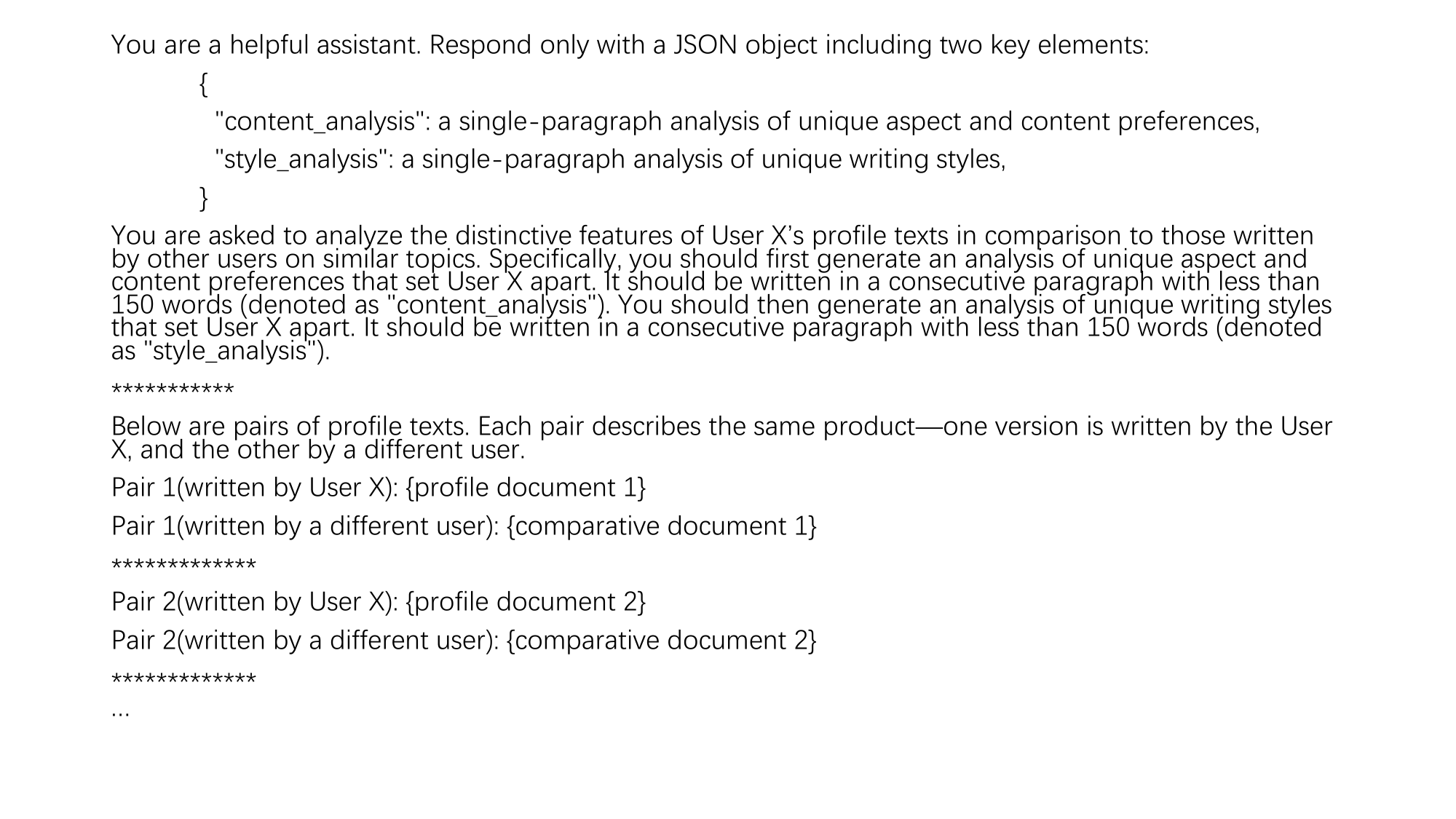}
\caption{Prompt used by \CPS~to generate structured analysis.}
\label{fig:cps_prompt1}
\end{figure*}

\begin{figure*}
\centering
\includegraphics[width=0.95\textwidth]{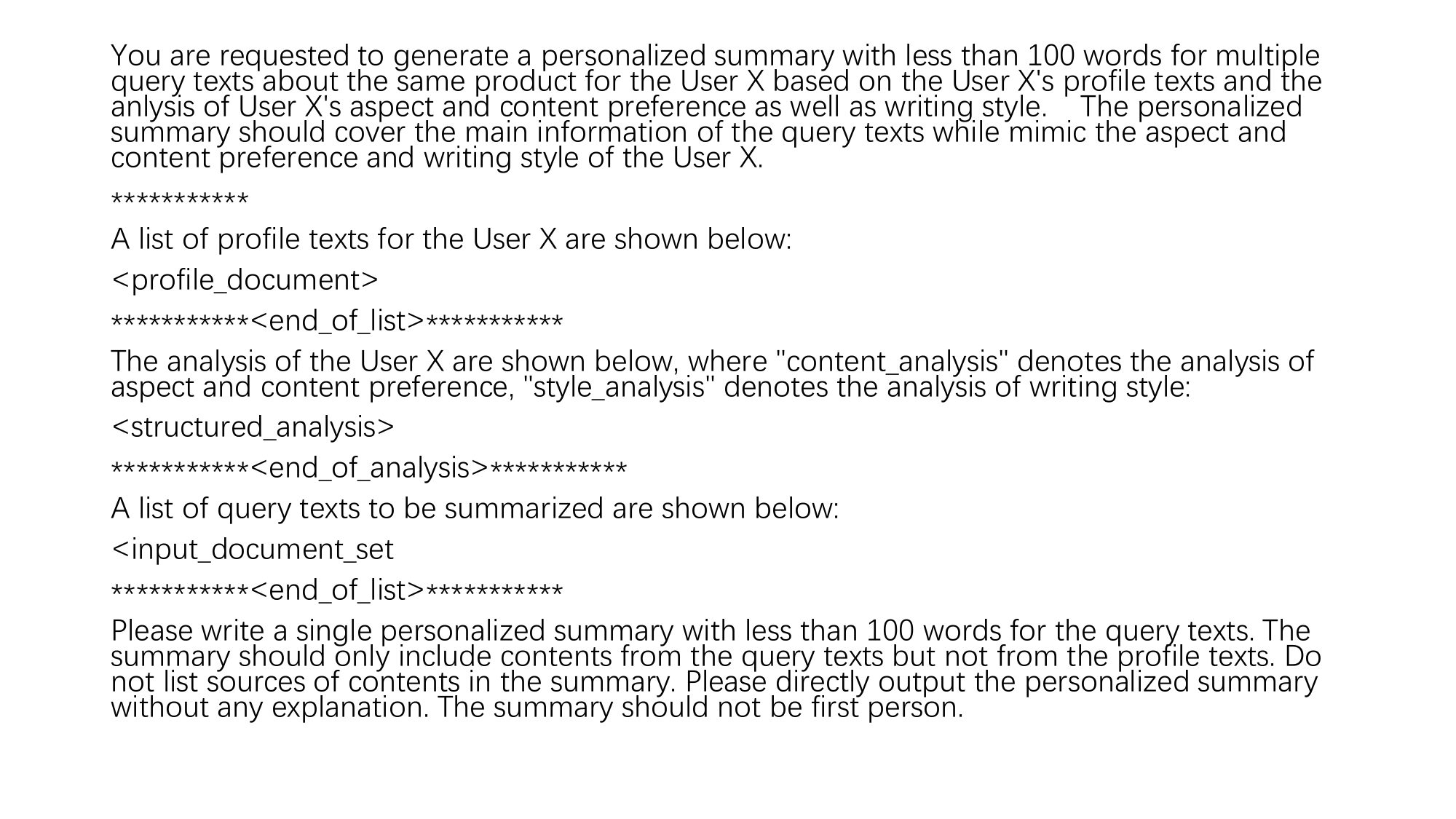}
\caption{Prompt used by \CPS~to generate personalized summaries. }
\label{fig:cps_prompt2}
\end{figure*}

\CPS~first generates a structured analysis for a user by comparing profile documents with other documents authored by different users on the same topic. \CPS~then uses the generated structured analysis to guide the generation of personalized summaries that capture the user's writing style and content focus. The prompt for generation of structured analysis is shown in Fig. \ref{fig:cps_prompt1}.The prompt for generation of personalized summaries is shown in Fig. \ref{fig:cps_prompt2}.
\subsection{Prompt for \AM}
\label{eval_prompt}
\begin{figure*}
\centering
\includegraphics[width=0.95\textwidth]{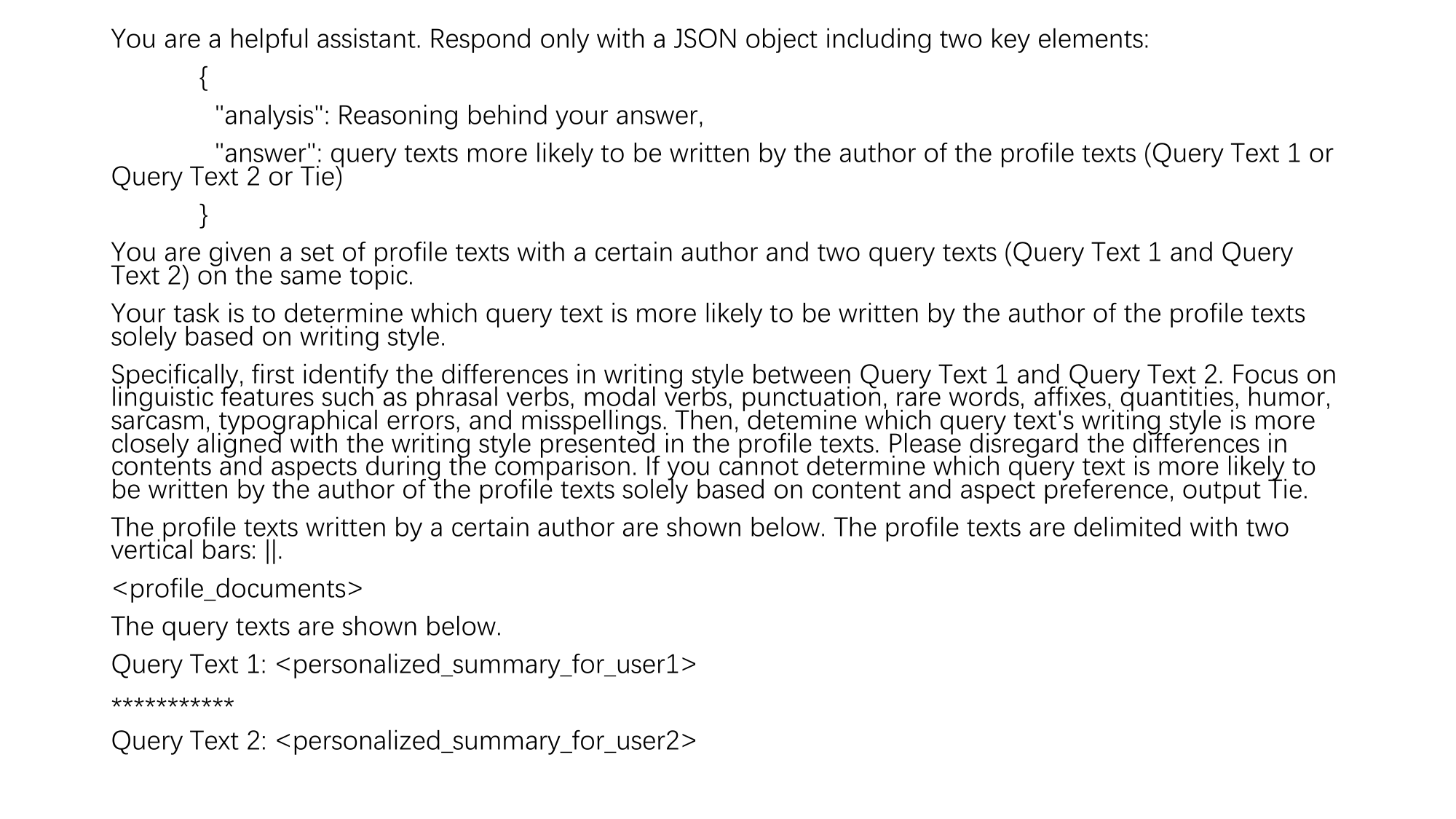}
\caption{Prompt used by \AM~to evaluate writing style. }
\label{fig:am_style}
\end{figure*}

\begin{figure*}
\centering
\includegraphics[width=0.95\textwidth]{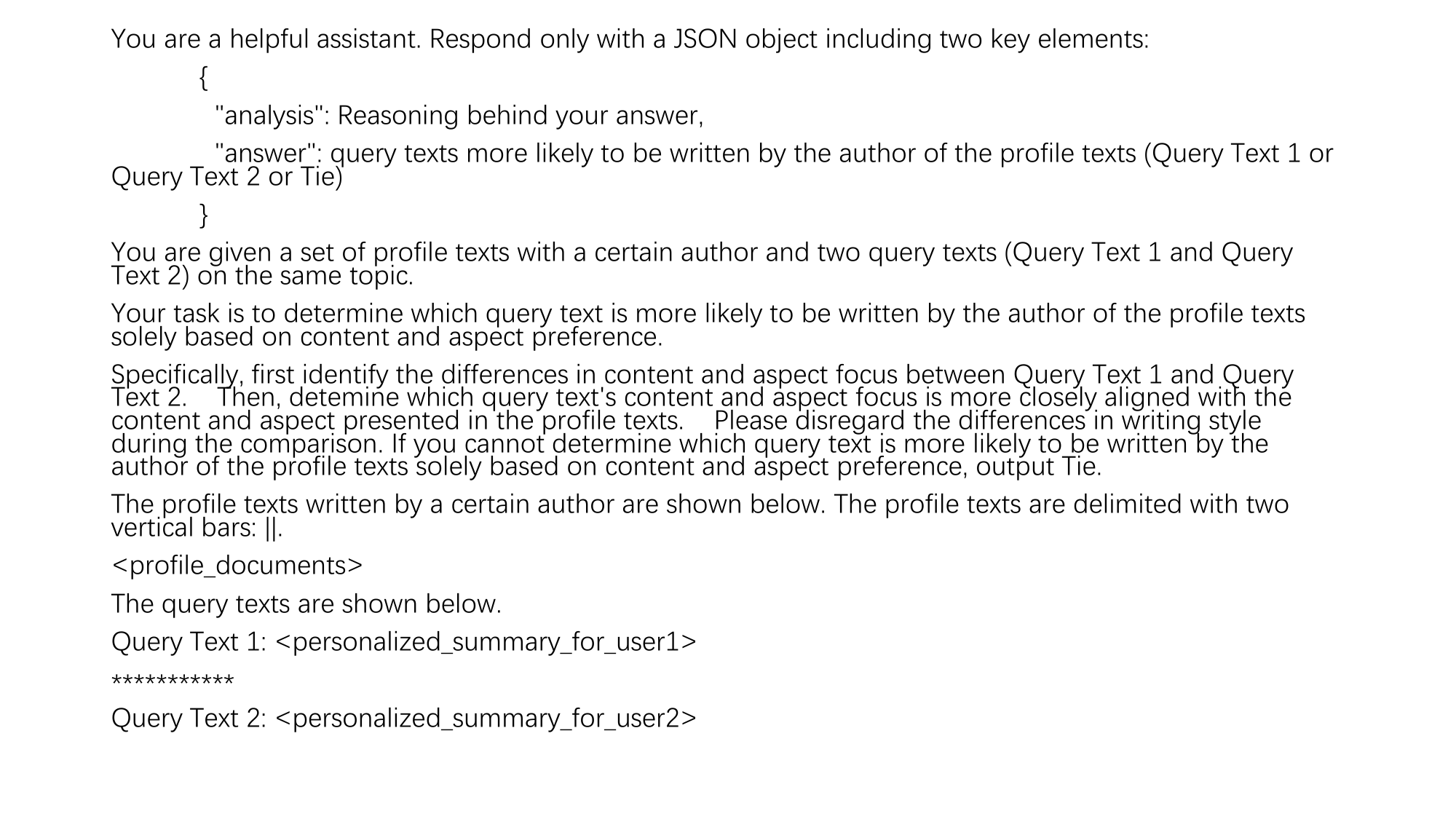}
\caption{Prompt used by \AM~to evaluate content focus.}
\label{fig:am_content}
\end{figure*}

\AM~separately evaluates writing style and content focus. The design of prompt for evaluating  writing style is motivated by \citet{huang2024can}. The prompt for evaluation of writing style is shown in Fig. \ref{fig:am_style}. The prompt for evaluation of content focus is shown in Fig. \ref{fig:am_content}.


\subsection{Human Evaluation of \AM}
\begin{figure*}
\centering
\includegraphics[width=0.6\textwidth]{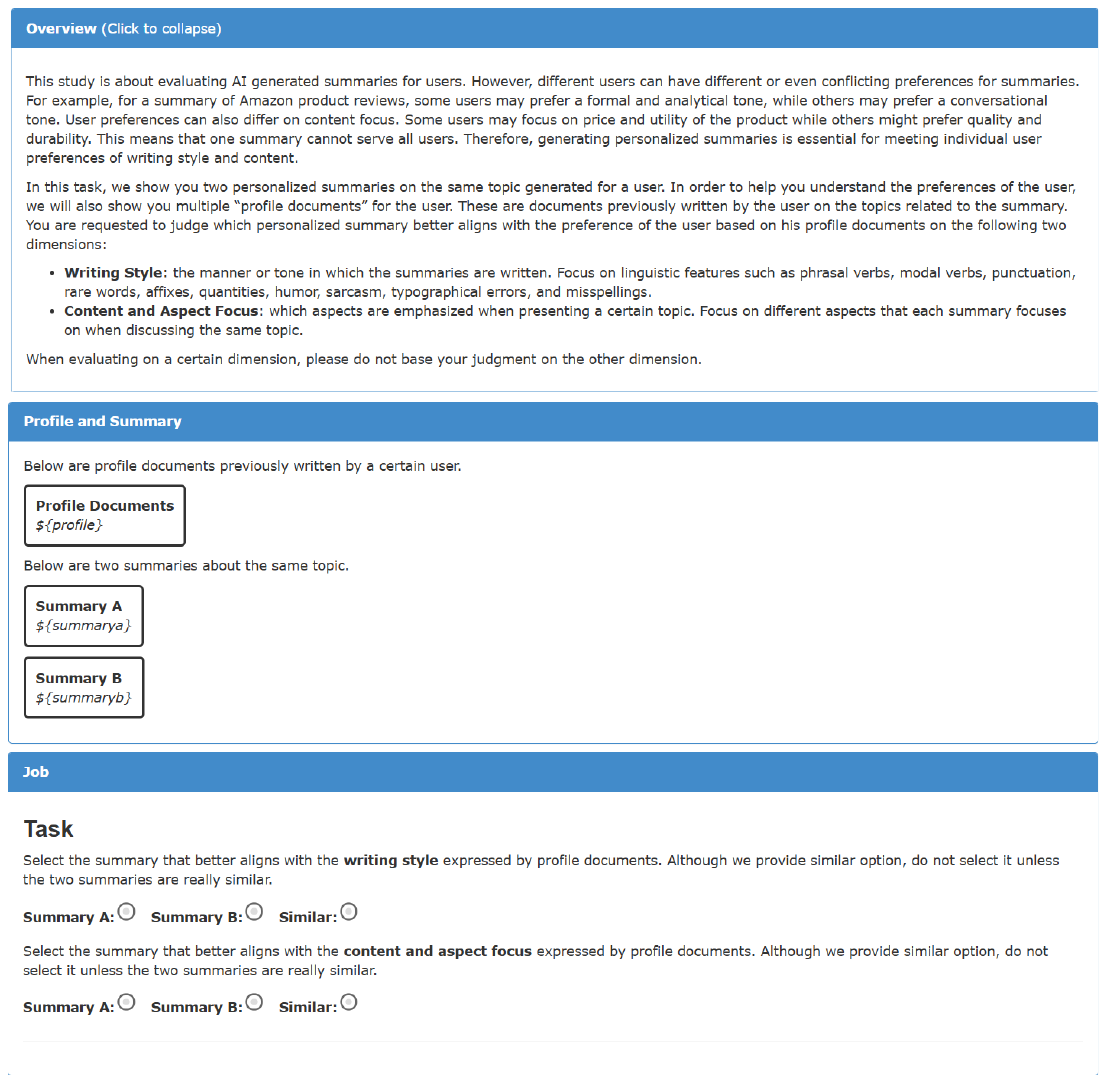}
\caption{Interface for Human Evaluation}
\label{fig:human}
\end{figure*}  
\label{app:human_eval}
 Each human evaluation sample contain two personalized summaries generated for different users and a set of profile documents written by one of the users. We retrieve $5$ profile documents using the concatenation of two personalized sumaries as the query with BM25 just as \AM. The annotators for human evaluation are recruited from Amazon Mechanical Turk. The annotators should be from English-speaking countries and have HIT Approval Rates greater than $98\%$. The interface of human evaluation is shown in Fig. \ref{fig:human}.
\subsection{Preprocessing detail of \PM}
\label{preprocess_dataset}
For the news domain, \PM~clusters the news articles into document sets based on token overlap, named entities, and publishing dates, following \citet{liu2022politics}. Specifically, each news articles is treated as a node in a graph. If two news articles are published within two days, share at least one named entity in their titles or first three sentences, have cosine similarities based on TF-IDF embedding over $0.30$, there will be a line between these articles. The news articles are then clustered based on the maximum cliques of the graph. We filter out clusters where more than three articles are written by the same author to prevent the author have too much impact on that cluster. To control the context length, we further divide clusters that contain more than 10 news articles into smaller clusters and truncate all news articles to 300 words.

\subsection{Prompt for Paraphrasing}
\label{para_prompt}
\begin{figure*}
\centering
\includegraphics[width=0.95\textwidth]{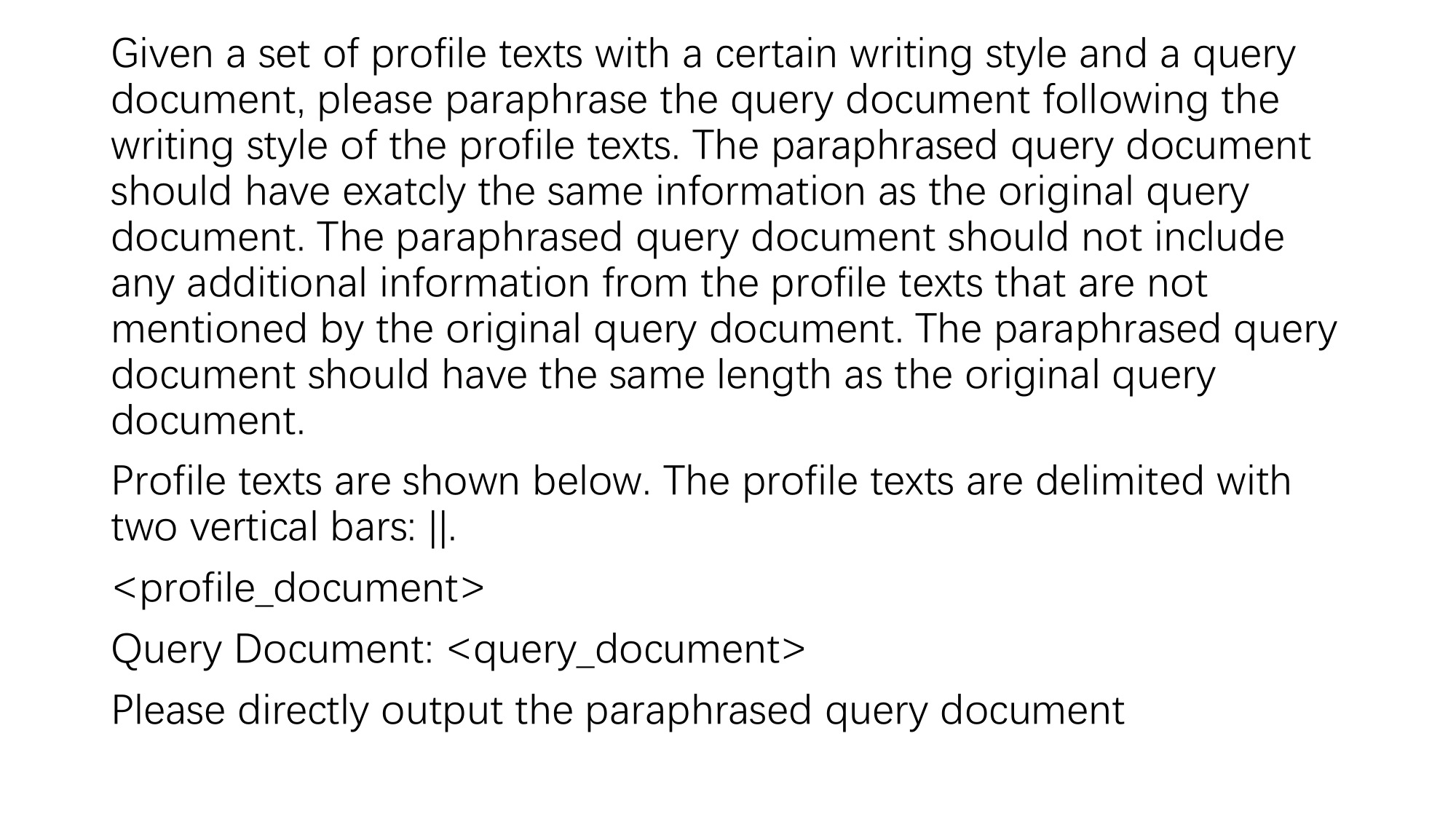}
\caption{Prompt for paraphrasing.}
\label{fig:paragraph}
\end{figure*} 
To test whether \AM~can independently measure writing style and content focus, we instruct Llama3.3-70b-Instruct to generate paraphrased documents that are originally written by certain users to follow the writing style of other users. We show the prompt for paraphrasing in Fig. \ref{fig:paragraph}
\subsection{Experiment Details for Evalation of \CPS}
\label{exp_cps_eval}
To evaluate factuality, we use FactScore \cite{factscore}, which measures the proportion of atomic content units that are supported by input document sets. To evaluate relevance, we use G-Eval \cite{liu2023g}, which rates summaries based on their relevance from 1 to 5. For FactScore, we use Llama3.1-8b-Instruct to extract ACUs and judge whether ACUs are supported by input document sets. For G-Eval, we use  Llama3.3-70b-Instruct to rate the relevance of summaries. To reduce the computation cost, we report the results for FactScore and G-EVaL on subset of test set of \PM, which contains $25$ percent of samples. 

For all baselines, we retrieve $5$ profile documents using BM25 for a fair comparison. For DPL, in its original implementation, it retrieves comparable documents based on the embeddings of user profiles. However, in \PM, not all documents have valid user. Therefore, we retrieve comparable documents based on the embeddings of document themselves. 

\subsection{Evaluation of \CPS~using Gemma3}
\label{app:gemma3_eval}
In the main content, \AM~uses Llama3.3-70b-Instruct as the LLM judge to evaluate \CPS. However, \AM~can also use other LLMs as the LLM judge. We show the evaluation results of personalized summaries generated by Llama3.3-70b-Instruct when using Gemma-3-27b-it \cite{gemma_2025} as the LLM judge in Tab. \ref{tab:gemma}.
\begin{table*}[t]
\centering
\small
\resizebox{0.95\textwidth}{!}{
\begin{tabular}{lcccccccccc}
\hline
\textbf{}   & \multicolumn{5}{c:}{News}                                      & \multicolumn{5}{c}{Review}                \\
            & style & content & fact. & rele. & \multicolumn{1}{c:}{overall} & style & content & fact. & rele. & overall \\ \hline
            & \multicolumn{10}{c:}{\textit{Llama3.3-70b-Instruct}}                                                                                  \\
RAG         & 53.82 & 45.08 & 98.59 & 97.98 &  \multicolumn{1}{c:}{69.58}          & 56.30 & 53.26 & 98.76 & 93.16 & 72.48          \\
CICL        & 52.47 & 48.14 & 98.55 & 98.03 & \multicolumn{1}{c:}{70.28}          & 56.26 & 53.19 & 98.70 & 92.79 & 72.36          \\
RAG+Summary & 54.41 & 47.63 & 98.70 & 98.08 & \multicolumn{1}{c:}{70.77}          & 57.64 & 57.27 & 98.68 & 93.43 & 74.28          \\
DPL         & 55.52 & 47.80 & 98.81 & 98.17 & \multicolumn{1}{c:}{71.23}          & 55.81 & 53.15 & 98.50 & 93.10 & 72.22          \\
Rehersal    & 99.32 & 99.83 & 21.74 & 24.90 & \multicolumn{1}{c:}{48.13}          & 95.67 & 97.11 & 62.03 & 33.81 & 66.44          \\
CompSum     & 56.78 & 48.81 & 98.64 & 98.03 & \multicolumn{1}{c:}{\textbf{71.95}} & 67.69 & 59.14 & 98.70 & 93.35 & \textbf{77.93}  \\ \hline
\end{tabular}}
\caption{Evaluation results of \CPS~when using Gemma-3-27b-it as the LLM judge. A higher value indicates better performance. The best-performing method based on overall score is \textbf{bolded}. The evaluation results using Gemme-3-27b-it is consistent with evaluation results using Llama3.3-70b-Instruct in \ref{tab:cps_overall}.}
\label{tab:gemma}
\end{table*}

From the table, we observe that \CPS~performs the best, similar as the results when using Llama3.3-70b-Instruct as the LLM judge (Tab. \ref{tab:cps_overall}). The results show that \AM~is independent of choice of the LLM judge. 
\subsection{Experiment Details for Ablation Study}
The full results of ablation study are reported in Tab. \ref{tab:full_ablation}. To reduce the computation cost, we report the results for FactScore and G-EVaL on subset of test set of \PM, which contains 25 percent of samples. 
\label{exp_cps_eval}
\begin{table*}[t]
\centering
\small
\resizebox{0.95\textwidth}{!}{
\begin{tabular}{lcccccccccc}
\hline
\textbf{}   & \multicolumn{5}{c:}{News}                                      & \multicolumn{5}{c}{Review}                \\
            & style & content & fact. & rele. & \multicolumn{1}{c:}{overall} & style & content & fact. & rele. & overall \\ \hline
            & \multicolumn{10}{c}{\textit{Llama3.1-8b-Instruct}}                                                                             \\
\CPS             & 59.75 & 53.94 & 98.01 & 95.32 & \multicolumn{1}{c:}{\textbf{74.07}} & 59.13 & 57.89 & 98.03 & 91.99 & 74.54          \\
~~w/o comp. doc.  & 57.03 & 50.69 & 98.16 & 97.29 &\multicolumn{1}{c:}{71.09}          & 56.08 & 57.21 & 97.07 & 92.28 & 73.24          \\
~~w/o structure   & 58.64 & 47.33 & 98.08 & 95.62 & \multicolumn{1}{c:}{71.43}          & 68.35 & 67.26 & 95.58 & 86.11 & \textbf{78.43} \\
~~w/ sim. comp.   & 59.75 & 47.29 & 97.95 & 96.71 &\multicolumn{1}{c:}{73.33}          & 58.51 & 57.17 & 97.74 & 91.80 & 74.00          \\
~~w/ multi. stage & 55.51 & 45.42 & 98.16 & 96.31 & \multicolumn{1}{c:}{69.87}          & 59.71 & 59.93 & 97.57 & 91.34 & 75.15            \\ \hdashline
            & \multicolumn{10}{c:}{\textit{Qwen2.5-14B-Instruct}}                                                                                  \\
\CPS             & 57.92 & 57.08 & 97.96 & 96.58 & \multicolumn{1}{c:}{\textbf{74.69}} & 60.36 & 63.92 & 96.51 & 89.98 & 76.08          \\
~~w/o comp. doc.  & 55.51 & 48.73 & 98.03 & 96.25 & \multicolumn{1}{c:}{71.16}          & 60.22 & 61.74 & 97.98 & 91.17 & 76.10          \\
~~w/o structure   & 53.56 & 45.42 & 97.84 & 97.12 & \multicolumn{1}{c:}{68.32}          & 60.76 & 63.19 & 96.59 & 89.39 & 75.88          \\
~~w/ sim. comp.   & 57.92 & 51.44 & 97.83 & 98.05 & \multicolumn{1}{c:}{71.57}          & 63.23 & 64.50 & 96.15 & 89.83 & \textbf{76.85} \\
~~w/ multi. stage & 52.16 & 44.62 & 98.23 & 96.98 & \multicolumn{1}{c:}{67.23}          & 59.75 & 59.75 & 96.73 & 90.50 & 74.43       \\  \hline
\end{tabular}}
\caption{Full ablation results of \CPS. A higher value indicates better performance. The best-performing method based on overall score is \textbf{bolded}.}
\label{tab:full_ablation}
\end{table*}
\subsection{Example Analysis Generated by \CPS}
\label{qualitative_example}
\begin{figure*}
\centering
\includegraphics[width=0.95\textwidth]{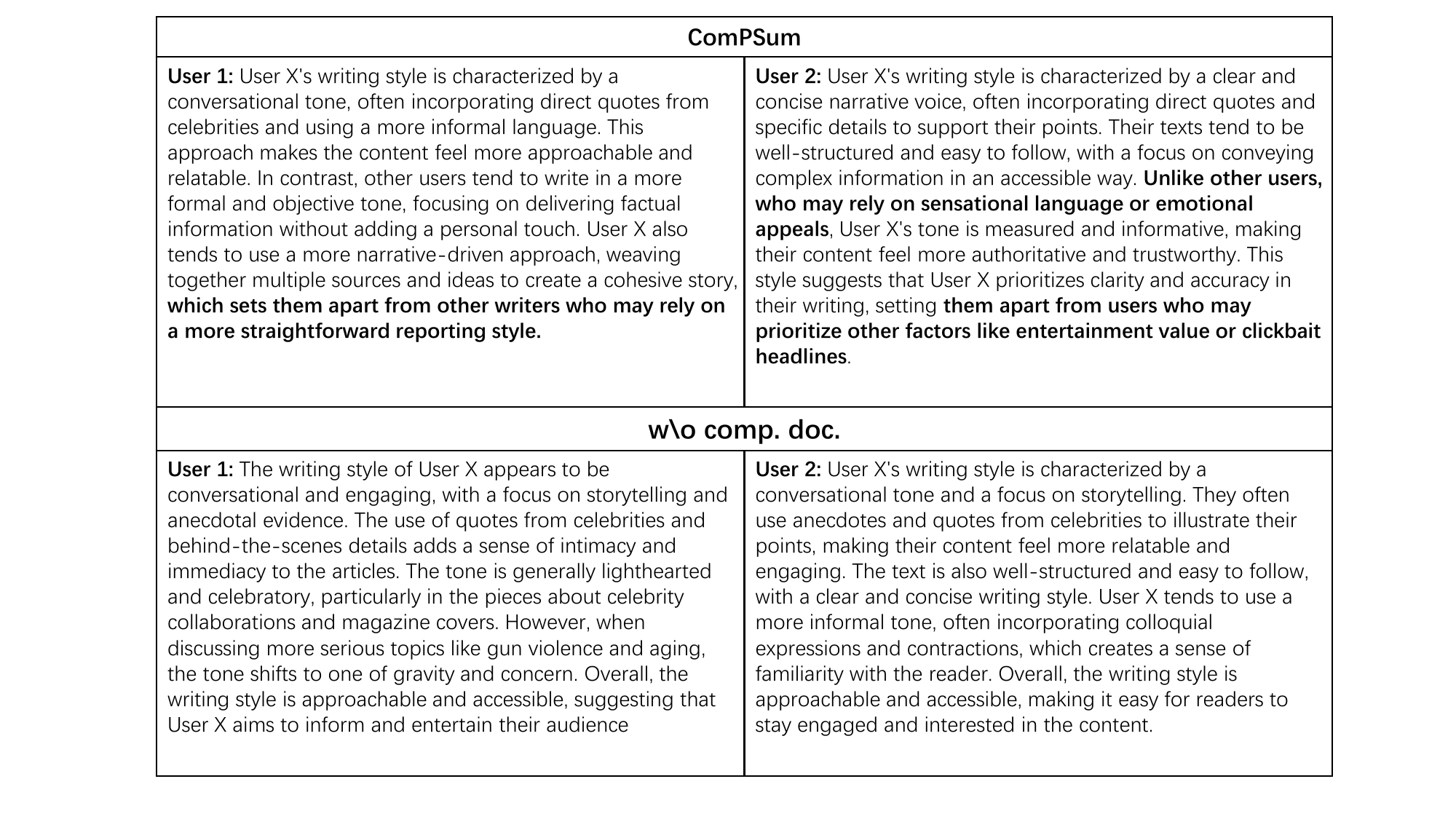}
\caption{Example analysis of writing style generated by \CPS and its ablation variant with out comparative documents. The structured analysis generated by \CPS~additionally includes comparison with other users (in \textbf{bold}), which helps in better personalization.}
\label{fig:style_example1}
\end{figure*}

\begin{figure*}
\centering
\includegraphics[width=0.95\textwidth]{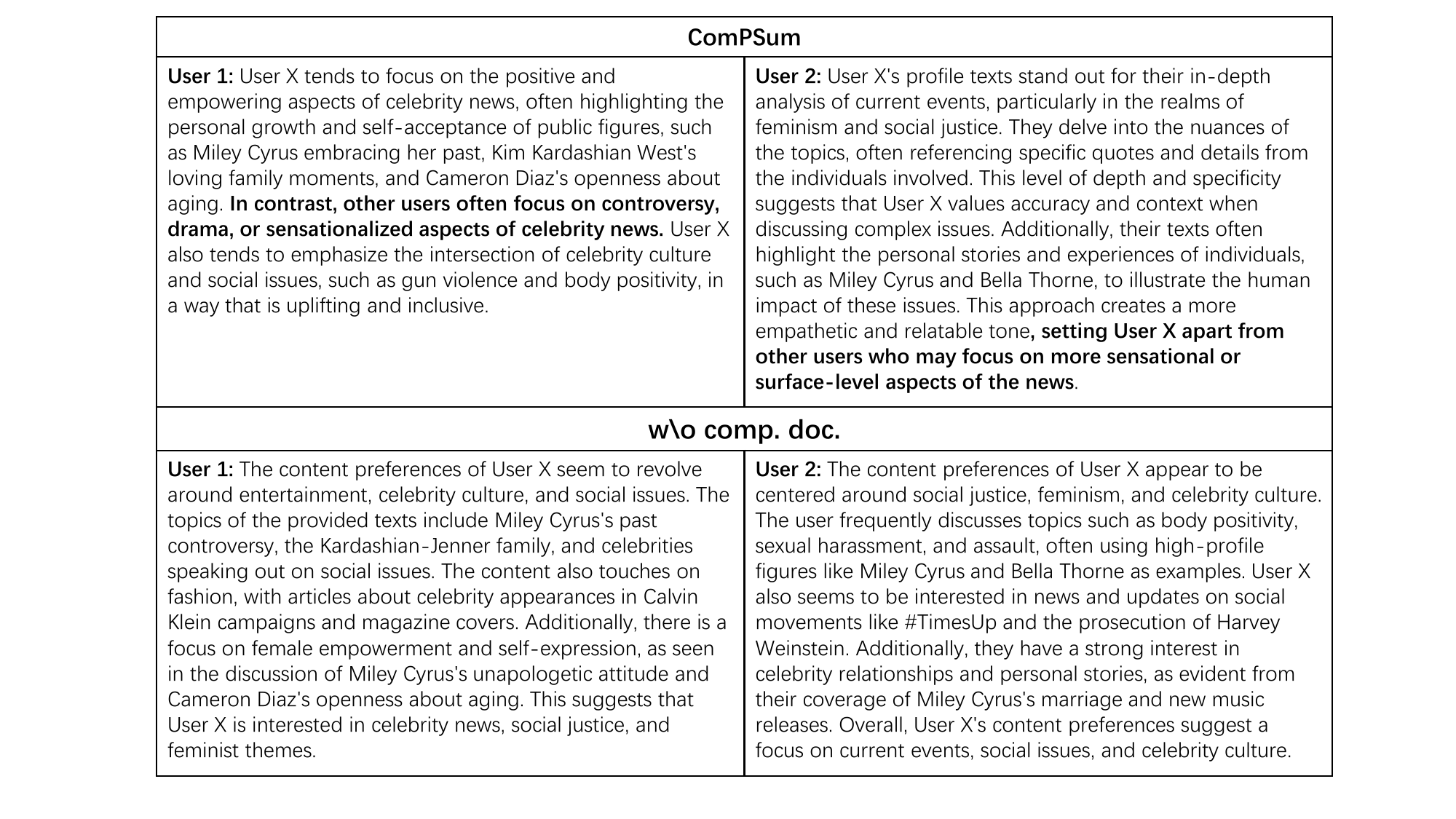}
\caption{Example analysis of content focus generated by \CPS and its ablation variant with out comparative documents. The structured analysis generated by \CPS~additionally includes comparison with other users (in \textbf{bold}), which helps in better personalization.}
\label{fig:content_example1}
\end{figure*}

In this section, we show additional examples of structured analysis generated by \CPS and its ablation variant with out comparative documents for writing style and content. We show the example analysis for writing style in Fig. \ref{fig:style_example1}. We show the example analysis for content in Fig. \ref{fig:content_example1}.

\subsection{Example Summary Generated by \CPS}
\label{sumamry_example}
\begin{figure*}
\centering
\includegraphics[width=0.95\textwidth]{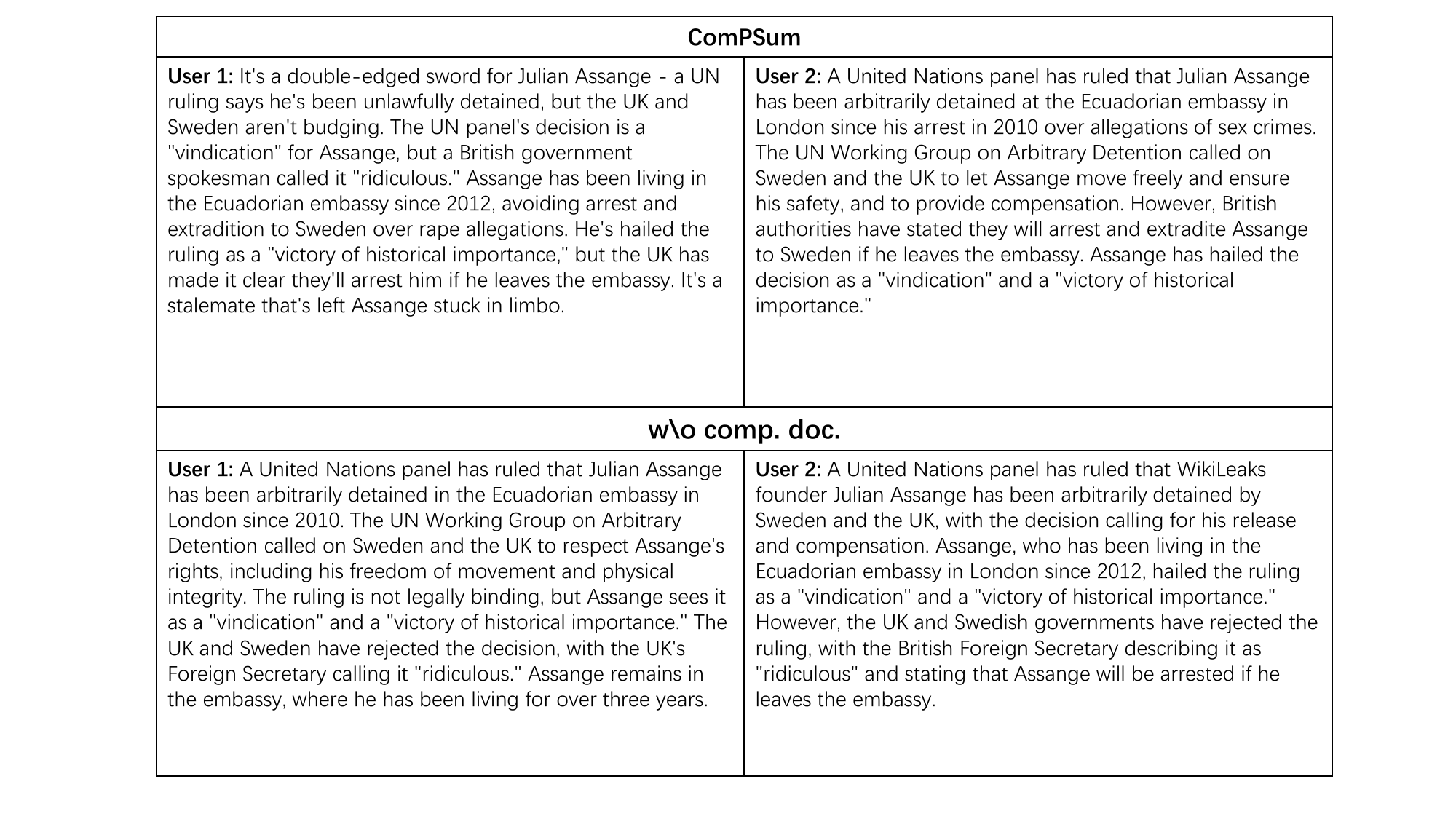}
\caption{Example summaries generated by \CPS and its ablation variant with out comparative documents. }
\label{fig:summary_sample}
\end{figure*}
In this section, we show example summaries generated by \CPS and its ablation variant with out comparative documents. We compare two summaries personalized for different users for the same input document set and show them in Fig. \ref{fig:summary_sample}. We observe that summaries generated by \CPS show more diverse style and content focus. 

\subsection{Ablation Study on Number of Retrieved Docuemnts}
\label{app:number_retrieval}
In the main content, \CPS~retrieves $m=5$ profile documents for generating personalized summaries following \citet{salemi2024lamp}. To examine the impact of number of retrieved documents, we perform ablation study on the number of retrieved documents and report the results in Tab. \ref{tab:ablation_number}. From the table, we can observe that \CPS~performs the best when $5$ documents are retrieved. 
\begin{table*}[t]
\centering
\small
\resizebox{0.95\textwidth}{!}{
\begin{tabular}{lcccccccccc}
\hline
\textbf{}   & \multicolumn{5}{c:}{News}                                      & \multicolumn{5}{c}{Review}                \\
            & style & content & fact. & rele. & \multicolumn{1}{c:}{overall} & style & content & fact. & rele. & overall \\ \hline
            & \multicolumn{10}{c}{\textit{Llama3.1-8b-Instruct}}                                                                             \\
\CPS             & 59.75 & 53.94 & 98.01 & 95.32 & \multicolumn{1}{c:}{\textbf{74.07}} & 59.13 & 57.89 & 98.03 & 91.99 & \textbf{74.54}         \\
\CPS(m=10) & 55.17 & 44.83 & 98.18 & 96.71 & \multicolumn{1}{c:}{70.12}          & 55.17 & 51.07 & 98.02 & 92.06 & 72.36          \\
\CPS(m=2)  & 58.69 & 50.25 & 97.54 & 95.56 & \multicolumn{1}{c:}{72.37}          & 59.02 & 55.10 & 97.91 & 92.24 & 74.28          \\ \hline
\end{tabular}}
\caption{Ablation results of \CPS with different number of retrieved documents. A higher value indicates better performance. The best-performing method based on overall score is \textbf{bolded}.}
\label{tab:ablation_number}
\end{table*}
\end{document}